\documentclass{ipol}

\ipolSetTitle{An Implementation of Two-Phase Image Segmentation 
              Using the Split Bregman Method}
\ipolSetAuthors{Olakunle Abawonse\ipolAuthorMark{1}, 
                G\"unay Do\u gan\ipolAuthorMark{2}}

\ipolSetAffiliations{
\ipolAuthorMark{1}African Institute of Mathematical Sciences (AIMS), Rwanda ({\tt olakunle.abawonse@aims.ac.rw})\\
\ipolAuthorMark{2}National Institute of Standards and Technology,
Gaithersburg, MD, USA ({\tt gunay.dogan@nist.gov}) }

\ipolSetEditors{
\emph{Communicated by} Enric Meinhardt-Llopis, Axel Davy, and Marina Gardella\qquad
\emph{Demo edited by} Marina Gardella
}
           
\ipolSetID{578}
\ipolSetVolume{16}

\ipolSetSubmissionYear{2024}
\ipolSetSubmissionMonth{09}
\ipolSetSubmissionDay{16}

\ipolSetAcceptedYear{2025}
\ipolSetAcceptedMonth{12}
\ipolSetAcceptedDay{26}

\ipolSetPublicationYear{2026}
\ipolSetPublicationMonth{03}
\ipolSetPublicationDay{00}

\usepackage[font={sf,footnotesize}]{caption}

\usepackage{hyperref,graphicx,amsmath,amssymb,amsthm}
\usepackage{url}

\usepackage{titlecaps}
\Addlcwords{a an on and for the of in to with from which as his their at or those by its et ses des de du la en les aux le dans zu}

\usepackage{enumerate}
\usepackage[vlined,ruled]{algorithm2e}
\SetKwInOut{Input}{input}
\SetKwInOut{Output}{output}

\newtheorem{definition}{Definition}
\newtheorem{theorem}{Theorem}

\DeclareMathOperator*{\argmin}{\arg\!\min}

\graphicspath{{Figures/}}

\begin{document}

\begin{ipolAbstract}
In this paper, we describe an implementation of the two-phase image segmentation algorithm proposed by Goldstein, Bresson and Osher in [Geometric Applications of the Split Bregman Method:
Segmentation and Surface Reconstruction, Journal of Scientific Computing, 2010]. This algorithm partitions the domain of a given 2D image into foreground and background regions, and each pixel of the image is assigned membership to one of these two regions. The underlying assumption for the segmentation model is that the pixel values of the input image can be summarized by two distinct average values, and that the region boundaries are smooth. Accordingly, the model is formulated as an energy functional whose variable is a region membership function that assigns pixels to either region, as originally proposed by Chan and Vese in [Active Contours Without Edges, IEEE Transactions on Image Processing, 2001]. This energy is the sum of image data terms in the regions and a length penalty for region boundaries. Goldstein, Bresson and Osher modify the energy of Chan-Vese so that their new energy can be minimized efficiently using the split Bregman method to produce an equivalent two-phase segmentation. We provide a detailed implementation of this method, and document its performance with several images over a range of algorithm parameters.
\end{ipolAbstract}
\begin{ipolCode}
The reviewed source code and documentation for this algorithm
are available at the  \href{\ipolLink}{IPOL web page of this article}. Compilation and usage instructions are included in the
\verb|README.txt| file of the archive.
\end{ipolCode}

\ipolKeywords{image segmentation; Bregman iterations; convex optimization}

\section{The Segmentation Model}\label{S:model}

Image segmentation is the problem of partitioning a given image into distinct regions, each of which is homogeneous with respect to image pixel values or image features. This problem is intrinsically hard as the image pixels often include noise or other local variability due to texture or other factors. A seminal mathematical model for achieving segmentation in such situations is the Mumford-Shah model proposed in~\cite{mumford:shah}. Given a gray-scale image $f:\Omega \rightarrow \mathbb{R}$ defined on the image domain $\Omega \subset \mathbb{R}^2$, a segmentation and smoothing are computed by minimizing the following model
\begin{equation}\label{eq:mumford-shah}
    E_{MS}(\Sigma,u)=\lambda_\Sigma|\Sigma| 
    +\int_\Omega (u(x)-f(x))^2 dx 
    + \mu\int_{\Omega-\Sigma}|\nabla u|^2 dx, \quad \mu,\lambda_\Sigma>0,
\end{equation}
with respect to the set of discontinuities or edges $\Sigma$ in the image, and a piecewise smooth approximation $u(x)$ to the image function $f(x)$. The minimizers $\Sigma^*$ and $u^*$ of~\eqref{eq:mumford-shah} are a segmentation and a smoothing/denoising of $f$ respectively. The edge length penalty $|\Sigma|$ in~\eqref{eq:mumford-shah} encourages smooth curves of discontinuity in the solution, and the integral of $|\nabla u|^2$ penalizes variations in $u$. The model~\eqref{eq:mumford-shah}, however, is hard to minimize computationally. 

A more tractable model can be obtained by assuming that the image $f$ can be approximated by a piecewise constant function instead of a piecewise smooth function, as observed in many images. This assumption is the basis of the Chan and Vese segmentation model~\cite{chan:vese}, which aims to segment the domain $\Omega \subset \mathbb{R}^2$ of the image $f$ into a foreground region $\Omega_1$ and a background region $\Omega_2 = \Omega \setminus \Omega_1$. The image values in $\Omega_1, \Omega_2$ can be approximated by averages $c_1, c_2$ respectively. Note, however, that more regions, or other image value models, such as piecewise affine linear functions, can also be assumed in~\eqref{eq:mumford-shah}~\cite{kiefer:sto}. The so-called two-phase segmentation problem of Chan and Vese~\cite{chan:vese,getreuer:CV} can be formulated as the minimization of the energy
\begin{equation}\label{eq:two-phase}
E_2(\Omega_1, \Omega_2, c_1, c_2) :=  \mbox{Per}(\Omega_1) 
+ \lambda \int_{\Omega_1} (f(x) - c_1)^2 dx 
+ \lambda \int_{\Omega_2} (f(x) - c_2)^2 dx,
\ \lambda > 0,
\end{equation}
where $\mbox{Per}(\Omega_1)$ is the perimeter of $\Omega_1$, i.e.\! the length of the boundary between $\Omega_1$ and $\Omega_2$. It is included in the energy~\eqref{eq:two-phase} as a regularization term to encourage smoothness of the region boundaries. The second and third terms in~\eqref{eq:two-phase} are the data fidelity terms to ensure that the regions $\Omega_1, \Omega_2$ match the perceived regions of the image, each approximated by the corresponding averages $c_1, c_2$ of pixel values. 
In the absence of the regularization term $\mbox{Per}(\Omega_1)$, the segmentation will be noisy, as is typically the case when the image pixels are thresholded into two sets, for example, by using the Otsu thresholding method for segmentation~\cite{otsu,balarini:nes}. 

The optimality of~\eqref{eq:two-phase} with respect to $c_1, c_2$ implies $c_1 = \frac{1}{|\Omega_1|} \int_{\Omega_1} f(x) dx, c_2 = \frac{1}{|\Omega_2|} \int_{\Omega_2} f(x) dx$. Thus the energy~\eqref{eq:two-phase} can be reduced to an energy in terms of $\Omega_1, \Omega_2$ only. This results in a region optimization problem. Chan and Vese~\cite{chan:vese} encode the regions in a level set function $\phi(x):\Omega \rightarrow \mathbb{R}$:
\[
\Omega_1 = \{x: \phi(x) < 0\}, \qquad \Omega_2 = \{x: \phi(x) > 0\},
\]
and rewrite the energy in terms of the level set function
\begin{equation}\label{eq:chan-vese}
E_{CV}(\phi) = \int_{\Omega}|\nabla H_{\epsilon}(\phi)| dx + \lambda \int_{\Omega} H_{\epsilon}(\phi) (f(x) - c_1)^2 + (1 - H_{\epsilon}(\phi))(f(x) - c_2)^2 dx,
\end{equation}
in which $H_{\epsilon}$ is a regularized Heaviside function. The first integral in~\eqref{eq:chan-vese} approximates the perimeter term in~\eqref{eq:two-phase}, and the second integral approximates the data fidelity terms in~\eqref{eq:two-phase}. A minimizer of~\eqref{eq:chan-vese} is computed by starting from a given initial level set function $\phi_0(x)$ and using the following time-dependent partial differential equation (PDE) to evolve $\phi(x)$,
\begin{equation}\label{eq:chan-vese-evolution}
\phi_t = H_{\epsilon}'(\phi) \left( \mbox{div} (\frac{\nabla \phi }{|\nabla \phi|} ) 
- \lambda((f(x) - c_1)^2 - (f(x)- c_2)^2)  \right).
\end{equation}

Equation~\eqref{eq:chan-vese-evolution} is a gradient descent evolution for the energy~\eqref{eq:chan-vese}. A stable and efficient numerical discretization and implementation of the PDE~\eqref{eq:chan-vese-evolution} requires care, and computing the solution can be slow. 

Chan, Esedoglu and Nikolova note in~\cite{chan:ese} that the following evolution equation
\begin{equation}\label{eq:u-pde-evolution}
    \frac{\partial u}{\partial t} =  \mbox{div} (\frac{\nabla u }{|\nabla u|} ) 
- \lambda((f(x) - c_1)^2 - (f(x)- c_2)^2), 
\end{equation}
has the same stationary points as Equation~\eqref{eq:chan-vese-evolution}, and it can be used to compute a solution $\Omega_1 = \{x: u(x) < \mu\}, \ \Omega_2 = \{x: u(x) > \mu\}$ ($\mu>0$ is some threshold) to the two-phase segmentation problem. Equation~\eqref{eq:u-pde-evolution} is the gradient descent evolution for the following energy
\begin{equation}\label{eq:convex-energy0}
E_0(u) = \|u\|_{TV(\Omega)} + 
\lambda \int_{\Omega}u(x)(f(x) - c_1)^2 + (1-u(x))(f(x) - c_2)^2 dx,
\quad \lambda>0,
\end{equation}
where $u(x) \in BV(\Omega)$ acts as a region indicator function, and $\|u\|_{TV(\Omega)}$, the total variation of $u$, is a regularization term, analogous to $\mbox{Per}(\Omega_1)$ in~\eqref{eq:two-phase}, to attain regions with smooth boundaries. The seminorm $\|u\|_{TV(\Omega)}$ is defined as
\begin{equation}\label{eq:TV-norm}
\|u\|_{TV(\Omega)}:= \int_{\Omega} |Du| := \sup\left\{\int_{\Omega} u\; \mbox{div}\, p\, dx : p \in C_c^1(\Omega, \mathbb{R}^2), |p(x)| \leq 1  \right\}.
\end{equation}

The total variation of a function $u$ was introduced in image processing by Rudin, Osher and Fatemi in~\cite{rudin:osher} to denoise images. For this, they evolved a noisy image $f$ using the total variation PDE, thereby reducing the total variation and noise content of $f$. In~\cite{cham:a}, Chambolle proposed a first order minimization algorithm for total variation minimization through nonlinear projections to convex sets in dual space. In~\cite{gstein:sh}, Goldstein et al.\! proposed to use Bregman iterations to solve total variation image denoising. The implementation of the algorithm from~\cite{gstein:sh} was described by Getreuer in~\cite{getr:p}. 

The following theorem connects the minimizer $u$ of the energy~\eqref{eq:convex-energy0} with the solution of the two-phase segmentation problem and the energy~\eqref{eq:two-phase}. Note that $\int_\Omega (f(x) - c_2)^2 dx$ in~\eqref{eq:convex-energy0} is constant for fixed $c_1, c_2$, and can be disregarded for this result.
\begin{theorem}\cite{chan:ese}
	For any given fixed $c_1, c_2 \in \mathbb{R}$, a global minimizer for energy~\eqref{eq:two-phase} is obtained by solving the convex minimization
\begin{equation}\label{eq:min-convex-energy0}
\min_{0 \leq u \leq 1} \int_{\Omega} |\nabla u| dx 
+ \lambda \int_{\Omega}\left\{ (f(x) - c_1)^2 - (f(x) - c_2)^2 \right\}u(x) dx,
\end{equation}
	 and then setting $\Omega_1 = \{x : u(x) \geq \mu\}$ for a.e.\! $\mu \in [0,1]$.
\end{theorem}
The above theorem suggests that we can take Equation~\eqref{eq:convex-energy0} (excluding the constant-valued $\int_\Omega (f(x) - c_2)^2 dx$ term) as the energy to be minimized and restrict our solution space to 
$\{u \in BV(\Omega): 0 \leq u(x) \leq 1\}$. 

Chan, Esedoglu and Nikolova observe in~\cite{chan:ese} that using a weighted total variation ($\mbox{TV}_g$) seminorm instead of $\|u\|_{TV(\Omega)}$ in~\eqref{eq:TV-norm} yields better segmentations as the $\mbox{TV}_g$ seminorm suppresses diffusion across region boundaries better. The weighted total variation seminorm $\|u\|_{TV_g(\Omega)}$ of $u$ is defined as
\begin{equation}\label{eq:TVg-norm}
\|u\|_{TV_g(\Omega)}:= \int_{\Omega} |Du|_g := \sup\left\{\int_{\Omega} u\; \mbox{div}\, p\, dx : p \in C_c^1(\Omega, \mathbb{R}^2), |p(x)| \leq g(x)  \right\}, 
\end{equation}
where $g(x)$ is a weight function defined using the image function $f(x)$. It is chosen so that it has a small value close to region boundaries or edges in the image, and a value close to 1 away from the edges. The following is a commonly used choice for $g(x)$, which we used in our implementation 
\begin{equation}\label{eq:weight-func}
    g(x) = 1 \, / \,  ( 1 + |(\nabla G_\sigma \ast f)(x)|^2 / \rho^2 ), \ \rho > 0,
\end{equation}
where $G_\sigma = \frac{1}{2\pi\sigma^2} \exp{\left(-\frac{|x|^2}{2\sigma^2}\right)}$ is the Gaussian convolution kernel with standard deviation $\sigma$, used to compute smoother derivatives of $f(x)$ to mark the edges in the image. Including weighted total variation~\eqref{eq:TVg-norm} as a regularization in~\eqref{eq:convex-energy0} and assuming known averages $c_1, c_2$ leads to the following energy that we minimize to obtain $u$
\begin{equation}\label{eq:convex-energy-g}
E_g(u) = \int_{\Omega} g(x)|\nabla u| dx + \lambda \int_{\Omega}  ((f(x) - c_1)^2 - (f(x) - c_2)^2 ) u(x) dx.
\end{equation}

Given the energy~\eqref{eq:convex-energy-g}, we can consider an alternating optimization approach for the more general case of unknown region averages $c_1, c_2$. We first estimate $c_1, c_2$ given $u$, then minimize~\eqref{eq:convex-energy-g} with respect to $u$, and continue to alternate between these two steps until convergence. In~\cite{gold:bre}, Goldstein et al.\! proposed the split Bregman method to solve the minimization problem for~\eqref{eq:convex-energy-g}, which we review in the following sections. Moreno et al.\! used the same approach for a four-phase segmentation in~\cite{more:sury}. 

\section{Minimization with Bregman Iterations}\label{S:bregman}

The minimizer $u$ of~\eqref{eq:convex-energy-g} can be computed using Bregman iterations~\cite{gold:bre}. Thus, in this section, we give a brief introduction to the Bregman method. A detailed explanation can be found in~\cite{osher:burg}. Suppose $J$ and $H$ are (possibly non-differentiable) convex functionals defined on a Hilbert space $X$. The Bregman iterations can be used to solve a convex minimization problem of the form
\begin{equation}\label{eq:general-convex-energy}
\min_{u \in B} J(u) + \lambda  H(u), \ \lambda > 0.
\end{equation}
In order to state the iterative minimization algorithm for~\eqref{eq:general-convex-energy}, we will need to introduce the notions of subgradient and Bregman distance.
\begin{definition}
Suppose $J: X \rightarrow \mathbb{R}$ is a convex function and $u \in X$. An element $p \in X^*$ is called a subgradient of $J$ at $v$ if for all $u \in X$
    $$J(u) - J(v) - \langle p, u - v \rangle \geq 0. $$
The set of all subgradients of $J$ at $v$ is called the subdifferential of $J$ at $v$, and it is denoted by $\partial J(v)$.
\end{definition}
\begin{definition}
Suppose $J : X \rightarrow \mathbb{R} $ is a convex function, $u,v \in X$ and $p \in \partial J(v)$. Then the Bregman distance between points $u$ and $v$ is defined by
$$D^p_J(u, v):= J(u) - J(v) - \langle p, u-v \rangle. $$
\end{definition}
We have the following basic distance-like properties of the Bregman distance: 
\begin{enumerate}[(i)]
\item $D^p_J(v,v) = 0$, 
\item $D^p_J(u,v) \geq 0$,  
\item $D^p_J(u,v) \geq D^p_J(w,v)$ for $w$ on the line segment from $u$ to $v$. 
\end{enumerate}

Property (iii) can be shown by setting $w = \alpha u + (1-\alpha)v$ on the line segment between $u$ and $v$. 
\begin{equation*}
\begin{aligned}
D^p_J(w,v) =  J(w) - J(v) & - \langle p, \alpha u + (1-\alpha)v - v \rangle
=  J(w) - J(v) - \alpha \langle p, (u-v) \rangle, \\[6pt]
\mbox{so that \ }
D^p_J(u,v) - D^p_J(w,v) 
& =  J(u) - J(w) - \langle p, u-v \rangle + \alpha \langle p, u-v \rangle \\
& \geq J(u) - (\alpha J(u) + (1-\alpha) J(v)) - (1-\alpha)\langle p, u-v \rangle\\
& =  (1-\alpha) (J(u) - J(v) - \langle p, u-v \rangle ) 
 =  (1- \alpha ) D^p_J(u,v) \geq 0. \qquad \qquad {}
\end{aligned}
\end{equation*}

The Bregman distance can be used as part of Bregman iterations to solve the minimization problem~\eqref{eq:general-convex-energy}. Osher et al.\! defined the Bregman iterations as follows in~\cite{osher:burg}
\begin{equation}
	u^{k+1} = \argmin_u D^{p^k}_J(u, u^k) + \lambda H(u) , \ p^k \in \partial J(u^k).
\end{equation}

When $H$ is differentiable, for $u^{k+1}$ minimizing $D^{p^k}_J(u, u^k) + \lambda H(u)$, we have $p^k - \lambda \nabla H(u^{k+1}) \in \partial J(u^{k+1})$, so we can take $p^{k+1} = p^k - \lambda \nabla H(u^{k+1})$. Thus, Bregman iterations for differentiable $H$ are summarized in Algorithm~\ref{alg:Bregman-iters}.

\begin{algorithm}[!htbp]
\caption{Bregman iterations for differentiable $H$}
\DontPrintSemicolon
\SetKwInOut{Input}{input}
\SetKwInOut{Output}{output}

\Input{$u^0$ -- initial estimate}
\Output{$(u^k, p^k)$ -- Bregman iterates}

Initialize $p^0 \in \partial J(u^0)$\\

\For{$k = 0,1,2,\dots$}{
    $u^{k+1} = \argmin_u \; D_J^{p^k}(u,u^k) + \lambda H(u)$\\
    $p^{k+1} = p^k - \lambda \nabla H(u^{k+1})$
}

\label{alg:Bregman-iters}
\end{algorithm}

\begin{theorem}\cite{osher:burg}
    Assume $J$ and $H$ are convex functionals defined on $BV(\Omega)$, and $H$ is differentiable.  Then we have
    $H(u_{k+1}) \leq H(u_k)$.
    Moreover, assume there exists a minimizer $u^* \in BV(\Omega)$ of $H(\cdot)$ with $J(u^*) < \infty$. Then 
    \[
    D_J^{p_k}(u^*, u_k) \leq D_J^{p_{k-1}}(u^*, u_{k-1}) \;\; \mbox{and} \;\;  H(u_k) \leq H(u^*) + \frac{1}{k}J(u^*).
    \]
    In particular, $u_k$ is a minimizing sequence of Algorithm~\ref{alg:Bregman-iters} under the above assumptions.
\end{theorem}
For the segmentation problem~\eqref{eq:convex-energy-g} that we will consider, we are interested in Bregman iterations in the case of linear constraints $A u = z$, corresponding to the case of $H(u) = \frac{1}{2} \|A u - z\|^2$, namely,
$$
\mbox{min}_u J(u) + \frac{\lambda}{2}\|Au - z\|^2,
$$
which can be solved using Algorithm~\ref{alg:Bregman-iters}  as follows
\begin{align*}
u^{k+1} & = \argmin_u D^{p^k}_J(u, u^k) + \frac{\lambda}{2}\|Au - z\|^2 
          = \argmin_u J(u) - \langle p^k , u-u^k \rangle + \frac{\lambda}{2} \|Au - z \|^2, \\
p^{k+1} & = p^k - \lambda A^T(Au^{k+1} - z).
\end{align*}

Existence of $b_k$ from $p_k = \lambda A^* b_k$ allows us to write a more compact form of update iterations.
The above iterations were shown in~\cite{osher:burg} to be equivalent to Algorithm~\ref{alg:Bregman-for-linear} when $A$ is linear.

\begin{algorithm}[!htb]
\caption{Bregman iterations for $H(u)=\frac{1}{2}\|Au - z\|^2$}
\DontPrintSemicolon
\SetKwInOut{Input}{input}
\SetKwInOut{Output}{output}

\Input{$z$ -- given data, $\lambda$ -- regularization parameter}
\Output{$(u^k, b^k)$ -- Bregman iterates}

Initialize $u^0$, $b^0 = 0$\\

\For{$k = 0,1,2,\dots$}{
    $u^{k+1} = \argmin_u \; J(u) + \frac{\lambda}{2}\|Au - z - b^k\|^2$\\
    $b^{k+1} = b^k + z - Au^{k}$
}

\label{alg:Bregman-for-linear}
\end{algorithm}

Now we describe how we build on Bregman iterations to minimize the segmentation energy~\eqref{eq:convex-energy-g}. This approach was proposed in~\cite{gold:bre}.
We first introduce an auxiliary variable $d:=\nabla u$ resulting in the constrained minimization problem
\begin{equation*}
\tilde{E}(u,d) = \int_{\Omega} g(x)|d| dx 
+ \lambda \int_{\Omega}  ((f(x) - c_1)^2 - (f(x) - c_2)^2 ) u(x) dx 
\qquad \mbox{s.t.} \quad d = \nabla u.
\end{equation*}
Then we write the equivalent augmented Lagrangian
\begin{equation}\label{eq:aug-Lagrangian}
\int_{\Omega} g(x)|d| dx + \lambda \int_{\Omega}  ((f(x) - c_1)^2 - (f(x) - c_2)^2 ) u(x) dx + \frac{\gamma}{2}\|d - \nabla u - b\|^2, \ \gamma > 0,
\end{equation}
which can be minimized with the customized Bregman iterations in Algorithm~\ref{alg:Bregman-segmentation0}.

\begin{algorithm}[!htb]
\caption{Bregman iterations for energy~\eqref{eq:aug-Lagrangian}}
\DontPrintSemicolon
\SetKwInOut{Input}{input}
\SetKwInOut{Output}{output}

\Input{$f$, $g$, $c_1$, $c_2$, $\lambda$, $\gamma$}
\Output{$(u^k, d^k, b^k)$ -- Bregman iterates}

Initialize $u^0 = 0$, $d^0 = 0$, $b^0 = 0$\\

\For{$k = 0,1,2,\dots$}{
    $(u^{k+1}, d^{k+1}) 
    = \underset{0 \leq u(x) \leq 1,\, d}{\operatorname{argmin}}
    \int_{\Omega} g(x)|d| 
    + \lambda \int_{\Omega} \big((f(x)-c_1)^2 - (f(x)-c_2)^2\big) u(x) 
    + \frac{\gamma}{2}\|d - \nabla u - b^k\|^2$\\
    
    $b^{k+1} = b^k + \nabla u^{k+1} - d^{k+1}$
}

\label{alg:Bregman-segmentation0}
\end{algorithm}

We will describe how to solve the minimization subproblem in Algorithm~\ref{alg:Bregman-segmentation0} in the next section.

\section{Discrete Model and Minimization}\label{S:discrete-model}

Although the image processing models and the Bregman minimization algorithm
to solve these models are expressed in function spaces, images are usually given
as 2D arrays of pixels. This necessitates working with discretized versions
of these models. Thus we start this section with a discretization of the differential
operators. Then we describe the discretized Bregman algorithms for two-phase image segmentation.

\subsection{Discrete Derivatives}\label{S:discrete-derivs}

In this section, we will define the discrete analogues of gradient, divergence and Laplacian operators on uniformly-spaced image grid $(u_{i,j}), (f_{i,j}), (g_{i,j}), (\vec{w}_{i,j}), i=0,\ldots,n_x, j=0,\ldots,n_y,$ of bounded functions $u, f, g, \vec{w}$.
The discrete first derivatives of $f_{i,j}$ are defined as
\begin{equation*}
\partial_x f_{i,j} := \left\{\begin{array}{cc}
f_{i+1,j} - f_{i,j}, & i = 0, \ldots, n_x-1, \\
0, & i = n_x,
\end{array}\right.
\quad
\partial_y f_{i,j} := \left\{\begin{array}{cc}
f_{i,j+1} - f_{i,j}, & j = 0, \ldots, n_y-1, \\
0, & j = n_y.
\end{array}\right.
\end{equation*}
The discrete gradient of $u_{i,j}$ is defined as
$
\nabla u_{i,j} := \displaystyle \left(\begin{array}{c}
\partial_x u_{i,j}\\
\partial_y u_{i,j}
\end{array}\right).
$ 
The discrete adjoint gradient of $g_{i,j}$ is 
$ -\nabla^* g_{i,j} := \displaystyle  \left(\begin{array}{c}
-\partial_x^* g_{i,j}\\  - \partial_y^* g_{i,j}
\end{array}\right), $ where
\[
- \partial_x^* g_{i,j} = 
\left\{\begin{array}{cc}
     g_{0,j},         & i = 0, \\
g_{i,j} - g_{i-1, j}, & 1 \leq i \leq n-1, \\
    -g_{n-1, j},      & i = n,
\end{array} \right.
\]
and $-\partial_y^* g_{i,j}$ is defined similarly.
For $\vec{w}_{i,j} = (w^x_{i,j}, w^y_{i,j})^T$, we have $\mbox{div}\; \vec{w}_{i,j} = -\partial_x^*w^x_{i,j} - \partial_y^*w^y_{i,j}$.
Finally, we define the discrete Laplacian as $\Delta u_{i,j} = -\partial_x^*\partial_x u_{i,j} - \partial_y^*\partial_y u_{i,j}$.

\subsection{The Discretized Segmentation Model}\label{S:discretized-segmentation}

In this section, we introduce the discretized version of the segmentation energy~\eqref{eq:convex-energy-g} for the given image data $\{f_{ij}\}$, and we describe the minimization algorithm for the discretized energy. We discretize the segmentation energy~\eqref{eq:convex-energy-g} as follows
\begin{equation}\label{eq:disc-convex-energy-g}
E(u)=\sum_{i,j} g_{i,j}|\nabla u_{i,j}| dx + \lambda \sum_{i,j}  ((f_{i,j} - c_1)^2 - (f_{i,j} - c_2)^2 ) u_{i,j}.
\end{equation}

As described in Section~\ref{S:discrete-model}, we introduce the auxiliary variable $d_{i,j} \approx \nabla u_{i,j}$, and consider the equivalent augmented Lagrangian. Then the key step in the Bregman iterations is the following minimization problem
\begin{equation}\label{eq:min-disc-aug-Lagrangian}
\argmin_{0 \leq u \leq 1, d} \sum_{i,j} g_{i,j} |d_{i,j}|+ \lambda \sum_{i,j} ((f_{i,j} - c_1)^2 - (f_{i,j} - c_2)^2) u_{i,j} +  \frac{\gamma}{2} \sum_{i,j} (d_{i,j} - \nabla u_{i,j} - b_{i,j})^2, 
\ \gamma > 0.
\end{equation}

To minimize the discretized augmented Lagrangian~\eqref{eq:min-disc-aug-Lagrangian}, we use the alternating direction method, namely, we minimize first over $d$, then over $u$ while keeping the other variable fixed. We iterate in this manner until convergence.

\paragraph{The $d$ subproblem:} With $u$ fixed, the $d$ subproblem is
\begin{equation}
\argmin_d \sum_{i,j} g_{i,j} |d_{i,j}| + \frac{\gamma}{2} \sum_{i,j} (d_{i,j} - \nabla u_{i,j} - b_{i,j})^2.
\end{equation}
The solution to the $d$ subproblem obtained in~\cite{gold:bre} is given by
\begin{equation}
d_{i,j} = \frac{\nabla u_{i,j} + b_{i,j}}{|\nabla u_{i,j} + b_{i,j}|} \max\{|\nabla u_{i,j} + b_{i,j}| - \frac{g_{i,j}}{\gamma} , 0\}.
\end{equation}

\paragraph{The $u$ subproblem:} With $d$ fixed, the $u$ subproblem is
\begin{equation}
\argmin_{0 \leq u \leq 1} \lambda \sum_{i,j} ((f_{i,j} - c_1)^2 - (f_{i,j} - c_2)^2) u_{i,j} + \frac{\gamma}{2} \sum_{i,j} (\nabla u_{i,j} - d_{i,j} + b_{i,j})^2.
\end{equation}
The optimal $u$ satisfies
\begin{equation}\label{eq:elliptic-pde0}
\lambda r_{i,j} + \gamma \nabla^* (\nabla u_{i,j} - d_{i,j} + b_{i,j}) = 0,
\end{equation}
where $r_{i,j} = (f_{i,j} - c_1)^2 - (f_{i,j}- c_2)^2$.
We rewrite~\eqref{eq:elliptic-pde0} as the following elliptic PDE
\begin{equation}\label{eq:elliptic-pde}
\Delta u_{i,j} = \frac{\lambda}{\gamma} r_{i,j} + \mbox{div}(d_{i,j}- b_{i,j}),
\end{equation}
which can be solved with a variety of PDE solvers, including Jacobi iterations. We do not, however, take Jacobi iterations to fully solve the PDE at each Bregman iteration. Instead, we execute an approximate solve with a single Jacobi iteration
$$
\beta^k_{i,j} = \frac{1}{4}(u^k_{i-1,j} + u^k_{i+1, j} + u^k_{i,j-1}+ u^k_{i,j+1} - \frac{\lambda}{\gamma} r_{i,j} + \mbox{div}(d_{i,j} - b_{i,j})) 
\  \mbox{for} \; 1 \leq i < n_x, 1 \leq j < n_y.
$$
Then we impose the bounds $0 \leq u \leq 1$ by
\begin{equation}
u^{k+1}_{i,j} = \max\{\min\{ \beta^k_{i,j}, 1\}, 0\}.
\end{equation}

\paragraph{Updating averages $c_1, c_2$:}
The region averages $c_1, c_2$ need to be recomputed at each iteration as the regions defined by $u_{i,j}$ change. In the discrete case, the regions are given by the sets of indices marking the spatial locations on the pixel grid
\begin{equation}\label{eq:disc-regions}
    \Omega_1^{k} = \{ (i,j): u_{i,j}^{k} \geq 0.5 \}, \qquad
    \Omega_2^{k} = \{ (i,j): u_{i,j}^{k} \leq 0.5 \}.
\end{equation}
The areas $ |\Omega_1^{k}|, |\Omega_2^{k}| $ of the two regions $\Omega_1^{k}, \Omega_2^{k}$ are the cardinalities of the sets~\eqref{eq:disc-regions}, namely, the counts of the indices contained in these two sets. Then the update formulas for $c_1^{k}, c_2^{k}$ are
\begin{equation}\label{eq:update-avgs}
c_1^{k} = \frac{1}{|\Omega_1^{k}|} \sum_{(i,j) \in \Omega_1^{k}} f_{i,j} , \qquad
c_2^{k} = \frac{1}{|\Omega_2^{k}|} \sum_{(i,j) \in \Omega_2^{k}} f_{i,j}.
\end{equation}

\subsection{More on Iterative Minimization}

Now we elaborate some other important pieces to implement an efficient and robust iterative algorithm for segmentation.
As in most iterative algorithms, we aim to ensure fast and consistent convergence to a minimizer of~\eqref{eq:min-disc-aug-Lagrangian}, or ideally of the energy~\eqref{eq:disc-convex-energy-g}. The value $E(u)$ serves as a measure of the quality of the segmentation encoded by $u$; the smaller$E(u)$, the better the segmentation. 
Accordingly, we track the energy values $E^k=E(u^k)$ through the iterations, continue as long as the $E^k$ values are decreasing, and stop if the decrease between iterations becomes too small.
Thus, our stopping criterion is 
\begin{equation}\label{eq:stopping-criterion}
    \| E^k - \bar{E}^{k-1} \| < \varepsilon_{tol} E^0,
\end{equation}
where $\bar{E}^{k-1}$ is the average of the previous $m$ energy values: 
$\bar{E}^{k-1} = \frac{1}{m}\sum^{k-1}_{l=k-m} E(u^l)$. We use the recent average $\bar{E}^{k-1}$ instead of $E^{k-1}$ to introduce some robustness to small fluctuations in the energy computation from noise in the image, and to avoid premature stopping of the iterations because of this. Since $\bar{E}^{k-1}$ needs $m$ energy values to compute the average, we always start with $m+1$ iterations (we set $m=10$ in our experiments).

We have also found it helpful to tune and dampen the updates $b^{k+1}=b^k+\nabla u^{k+1} - d^{k+1}$ in Algorithm~\ref{alg:Bregman-segmentation0}. A variable step size was employed in~\cite{goldste:li} to adjust the pace of the updates, and an upper bound was imposed on the step size in~\cite{goldstein:set} to ensure convergence of the algorithm. We found the use of a fixed step size $\tau$ to be practical and effective, and executed the updates to $b^{k+1}$ by
\begin{equation}\label{eq:b-update}
    b^{k+1} = b^k + \tau (\nabla u^{k+1}  - d^{k+1}).
\end{equation}
We experimented with various step sizes $\tau$, and picked a conservative value $\tau=0.01$. However, values in the range $[0.001,0.1]$ seemed to work well. Smaller step sizes resulted in more iterations, and larger step sizes led to a reduced number of iterations, but risked overshooting the minimum of the energy.

Finally, a good initialization is an important factor in achieving computational efficiency. Different works in the literature have used seeds or sets of shapes like circles as initialization~\cite{chan:vese}. A more effective strategy could involve a pre-clustering of the pixels based on their grayscale values into two clusters. In our experiments, we found the image itself to be a good initialization for the iterations. We normalized the grayscale image to the range $[f_{min},f_{max}]$, the minimum and maximum values of $f$ respectively, and initialized $u^0$ by
\begin{equation}\label{eq:init-u0}
    u^0 = (f - f_{min}) / (f_{max} - f_{min}).
\end{equation}
Incorporating the initialization, the stopping criterion and the step size gave us the final iterative algorithm for two-phase image segmentation, shown in Algorithm~\ref{alg:detailed-algorithm}.

\begin{algorithm}[!htb]
\caption{Minimization of discretized segmentation model}
\DontPrintSemicolon
\SetKwInOut{Input}{input}
\SetKwInOut{Output}{output}

\Input{$m$, $\varepsilon_{tol}$, $\tau$, $f$}
\Output{$u$}

Initialize $u^0$ by~\eqref{eq:init-u0}, 
$E^0 = E(u^0)$, 
$b^0 = 0$, 
$d^0 = 0$, 
$k=0$\\

\While{$(k<m)$ \textbf{or} $\big(\|E^k-\bar{E}^{k-1}\|>\varepsilon_{tol}E^0\big)$}{
    Solve the $u$ subproblem to obtain $u^{k+1}$\\
    Solve the $d$ subproblem to obtain $d^{k+1}$\\
    
    $\Omega_1^{k+1} = \{(i,j): u_{i,j}^{k+1} \geq 0.5\}$\\
    $\Omega_2^{k+1} = \{(i,j): u_{i,j}^{k+1} < 0.5\}$\\
    
    $c_1^{k+1} = \dfrac{1}{|\Omega_1^{k+1}|}
    \sum_{(i,j)\in\Omega_1^{k+1}} f_{i,j}$\\
    
    $c_2^{k+1} = \dfrac{1}{|\Omega_2^{k+1}|}
    \sum_{(i,j)\in\Omega_2^{k+1}} f_{i,j}$\\
    
    $b^{k+1} = b^k + \tau\big(\nabla u^{k+1} - d^{k+1}\big)$\\
    
    $E^{k+1} = E(u^{k+1})$\\
    $\bar{E}^{k} = \dfrac{1}{m}\sum_{l=k-m+1}^{k} E^l$\\
    
    $k = k+1$
}

\label{alg:detailed-algorithm}
\end{algorithm}

\section{Experiments}\label{S:experiments}

\begin{figure}[htb!]
	\begin{tabular}{ccc}
		\includegraphics[width=0.33\linewidth]{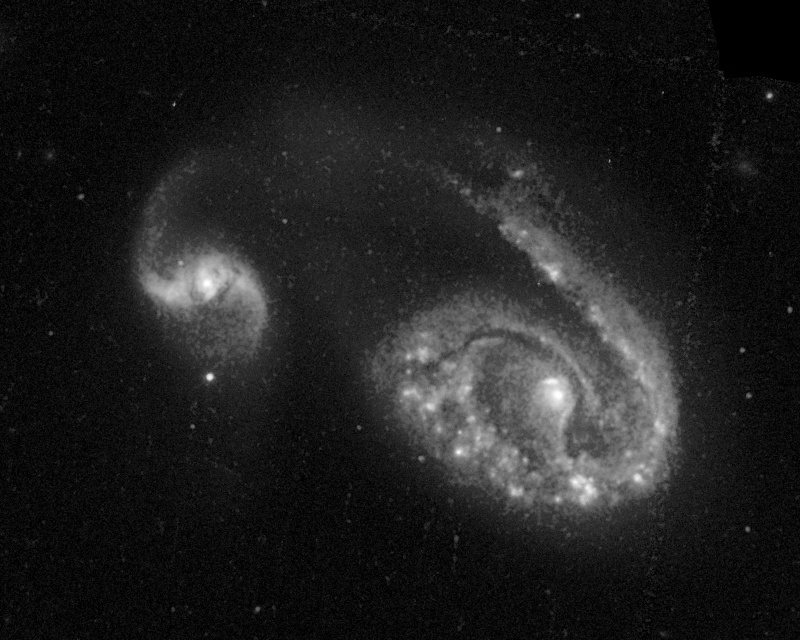} &
            \includegraphics[width=0.28\linewidth]{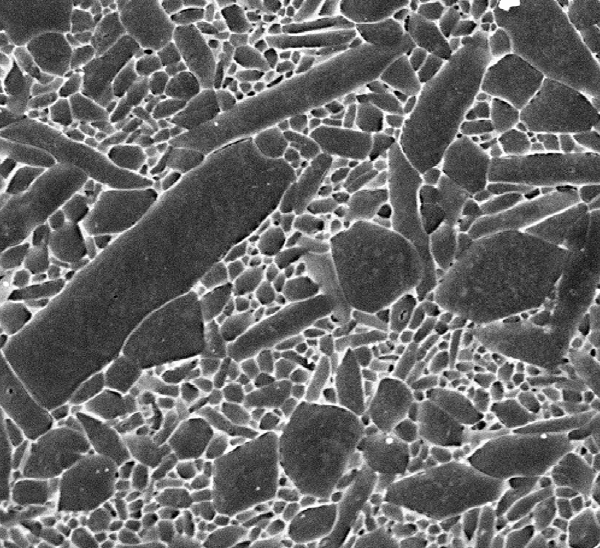} &
            \includegraphics[width=0.33\linewidth]{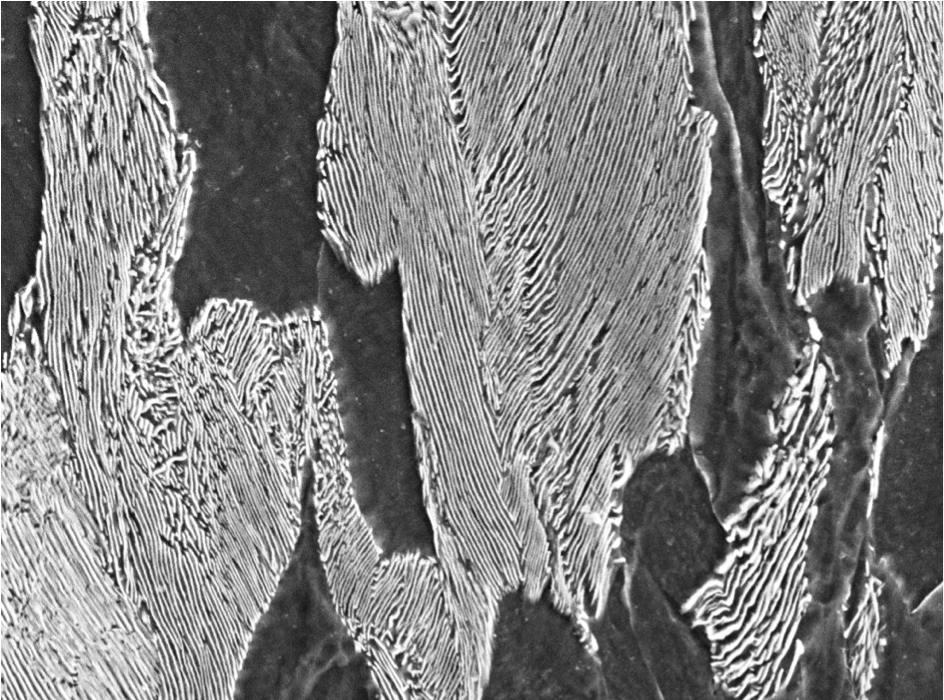} 
            \\
		{\sf \footnotesize galaxy, $ 640\times  800$} & 
		{\sf \footnotesize microstructure, $548\times 600$} &
		{\sf \footnotesize pearlite, $700\times 944$} 
	\end{tabular}
	\caption{Images used in the Experiments section}
	\label{fig:test-images}
\end{figure}

We implemented Algorithm~\ref{alg:detailed-algorithm} in Python, and tested it with experiments on real images. We used three test images: galaxy (image of two galaxies), microstructure (material microstructure image with voids), and pearlite (material microstructure: micrograph of etched pearlite), shown in Figure~\ref{fig:test-images}. Leveraging vectorized operations of the NumPy package in Python, we were able to code components of Algorithm~\ref{alg:detailed-algorithm} in a concise, easily-readable manner, which was also efficient for computation. We ran our experiments on a laptop computer with Intel\textsuperscript{\texttrademark} Core i7 10th Gen processor and 32GB of memory. We used the same algorithm parameters $\gamma=0.1, \tau=0.01, m=10, \varepsilon_{tol}=10^{-4}$ for all the experiments, but varied the model parameter $\lambda$ to illustrate its effect on the regularity of the solution. In the rest of this section, we describe the results of our experiments and our observations.

\clearpage

\begin{figure}[!htbp]
	\begin{tabular}{cccc}
		\includegraphics[width=0.23\linewidth]{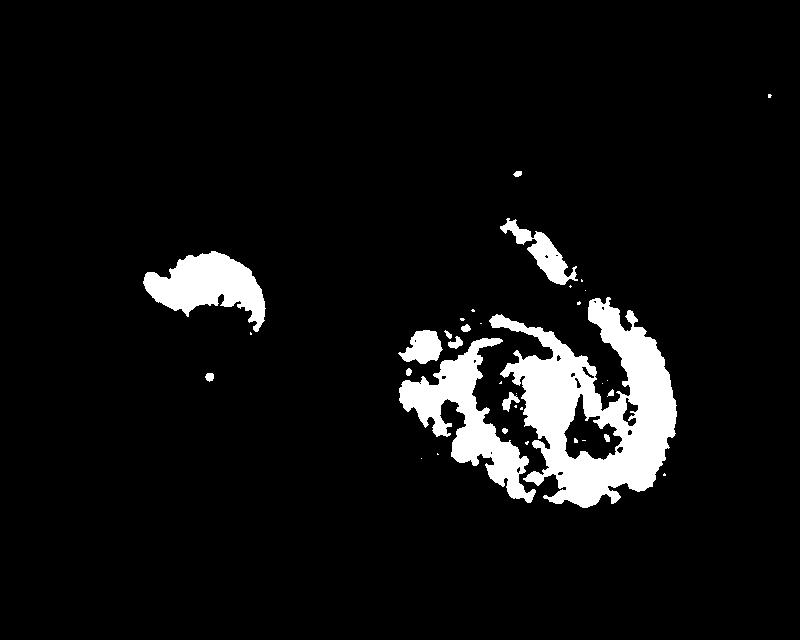} &
		\includegraphics[width=0.23\linewidth]{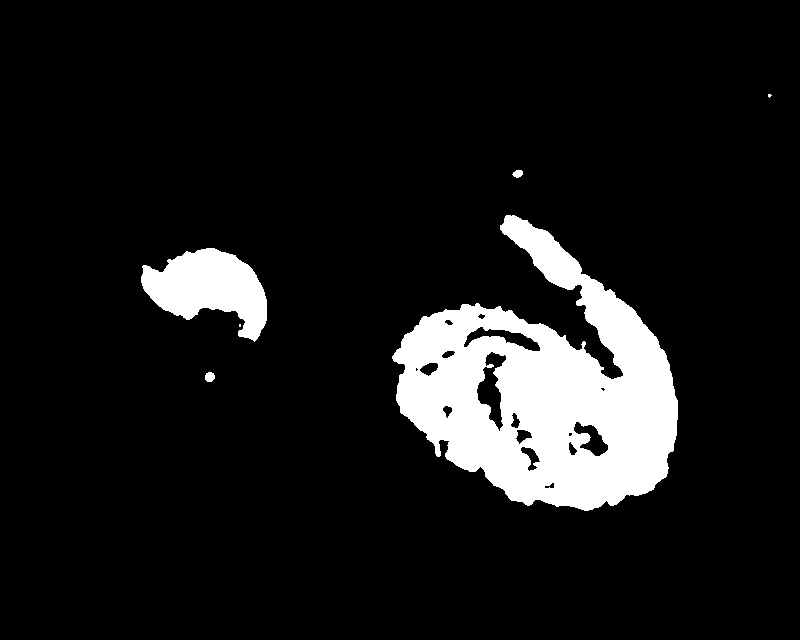} &
		\includegraphics[width=0.23\linewidth]{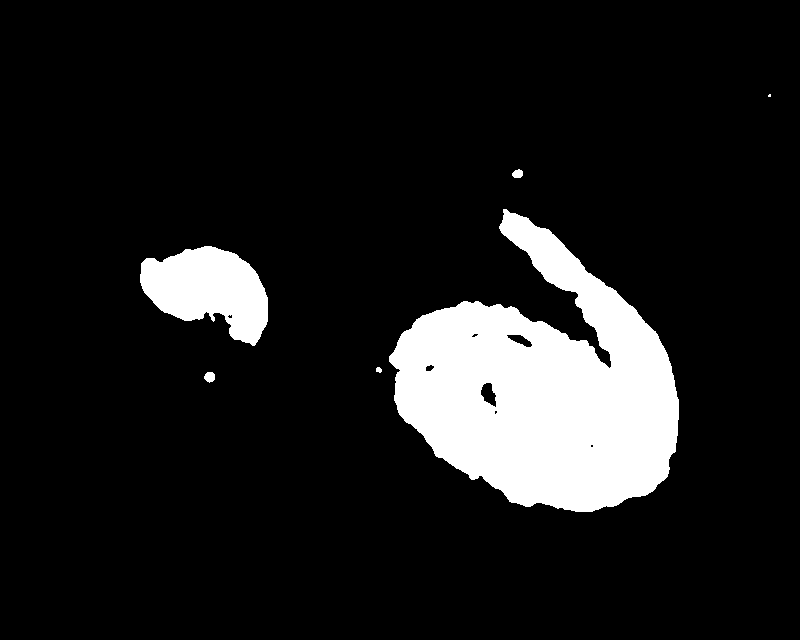} &  \includegraphics[width=0.23\linewidth]{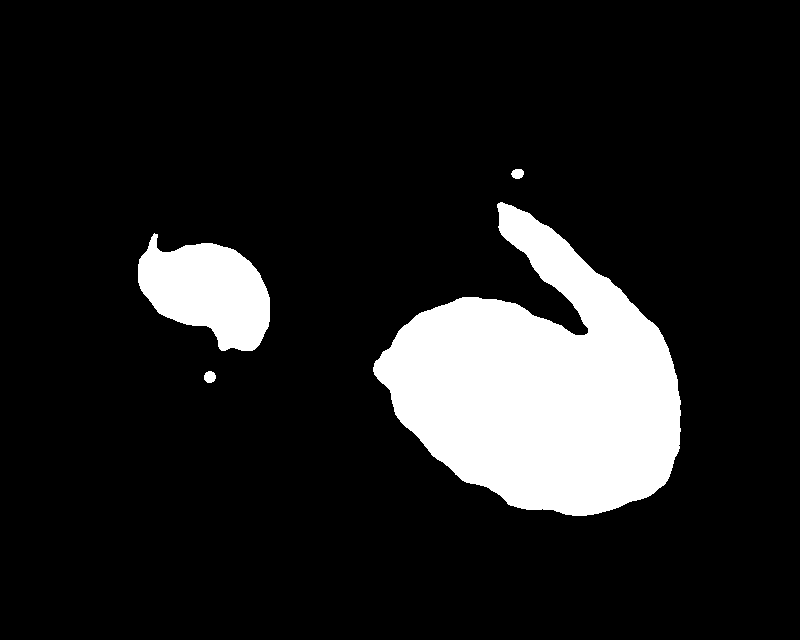}
		\\
		{\sf \footnotesize Iteration no: 5} & {\sf \footnotesize Iteration no: 20} & {\sf \footnotesize Iteration no: 40}  & {\sf \footnotesize Iteration no: 113}\\

		\includegraphics[width=0.23\linewidth]{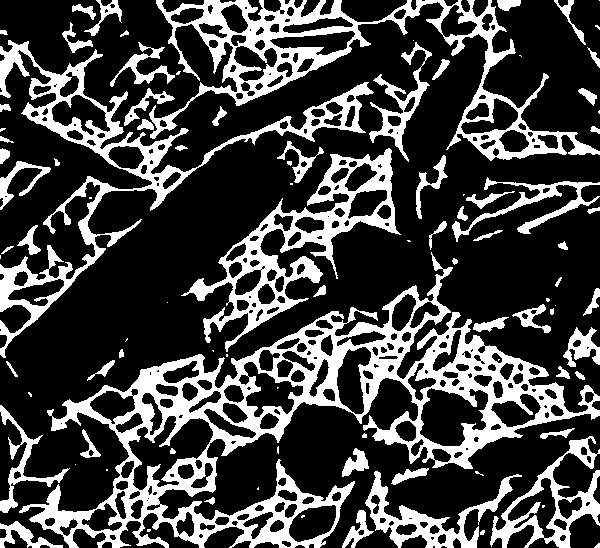} &
	    \includegraphics[width=0.23\linewidth]{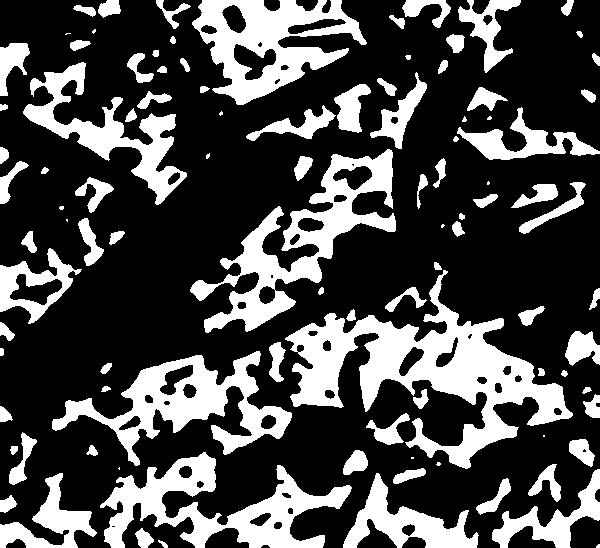}  &
	    \includegraphics[width=0.23\linewidth]{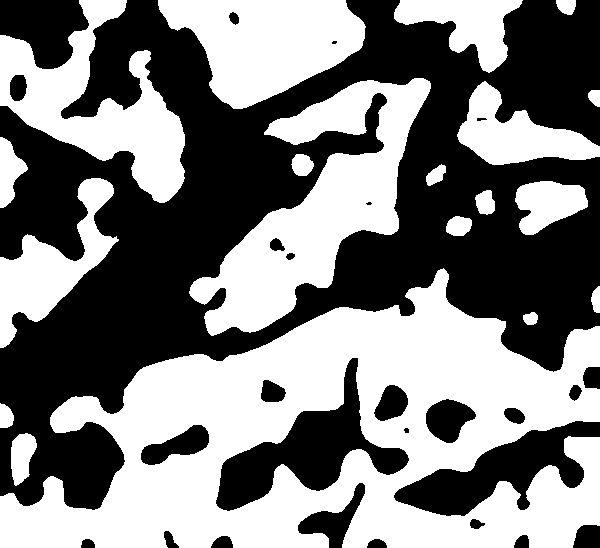} &  \includegraphics[width=0.23\linewidth]{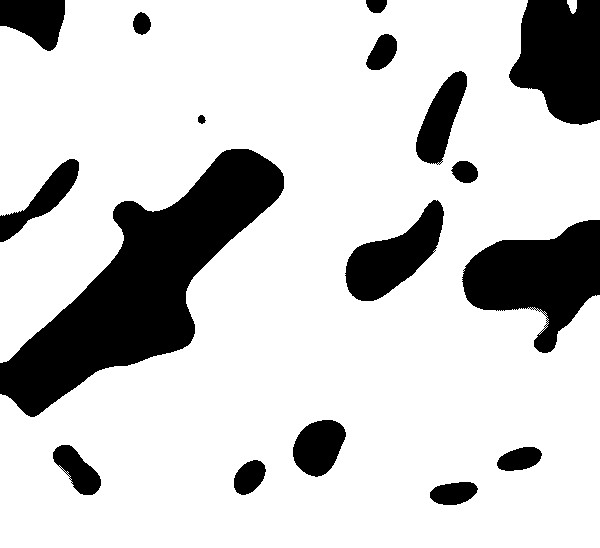}
	    \\
		{\sf \footnotesize Iteration no: 3} & {\sf \footnotesize Iteration no: 100} & {\sf \footnotesize Iteration no: 300}  & {\sf \footnotesize Iteration no: 853}\\

    	\includegraphics[width=0.23\linewidth]{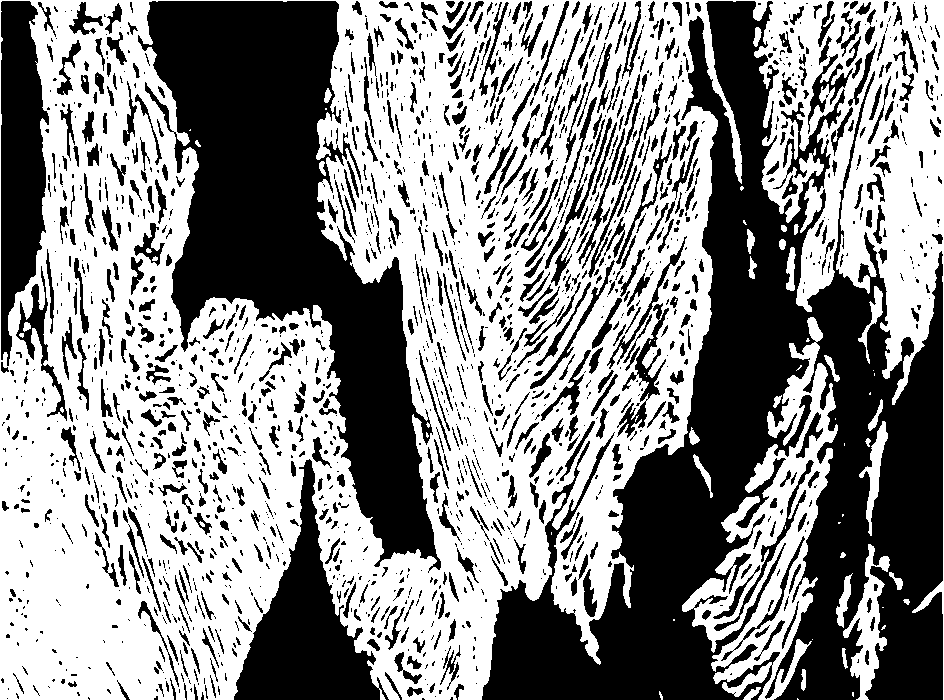} &
		\includegraphics[width=0.23\linewidth]{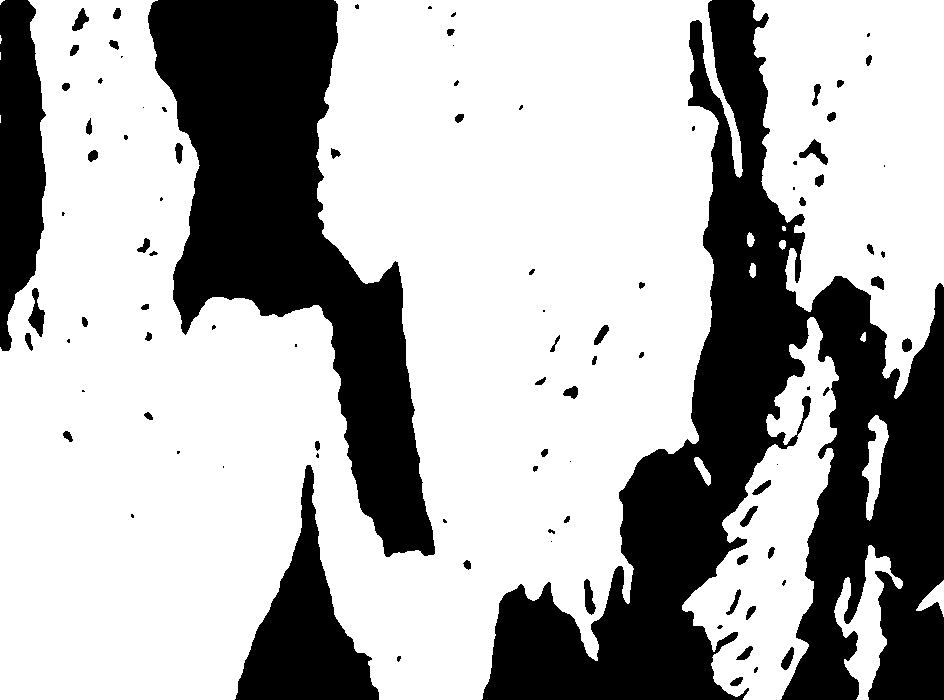}  &
		\includegraphics[width=0.23\linewidth]{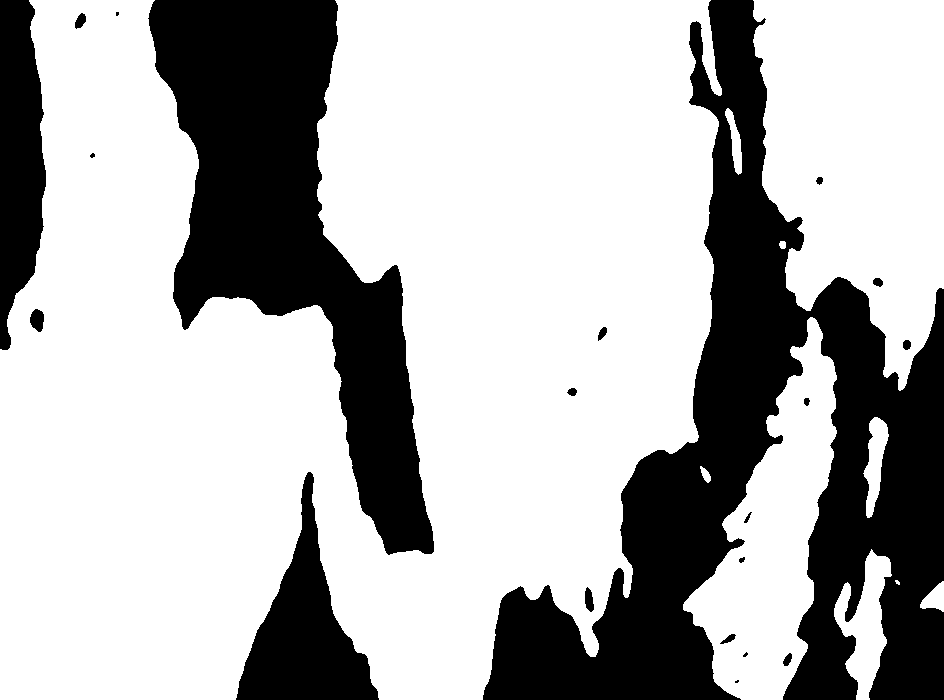} &   
        \includegraphics[width=0.23\linewidth]{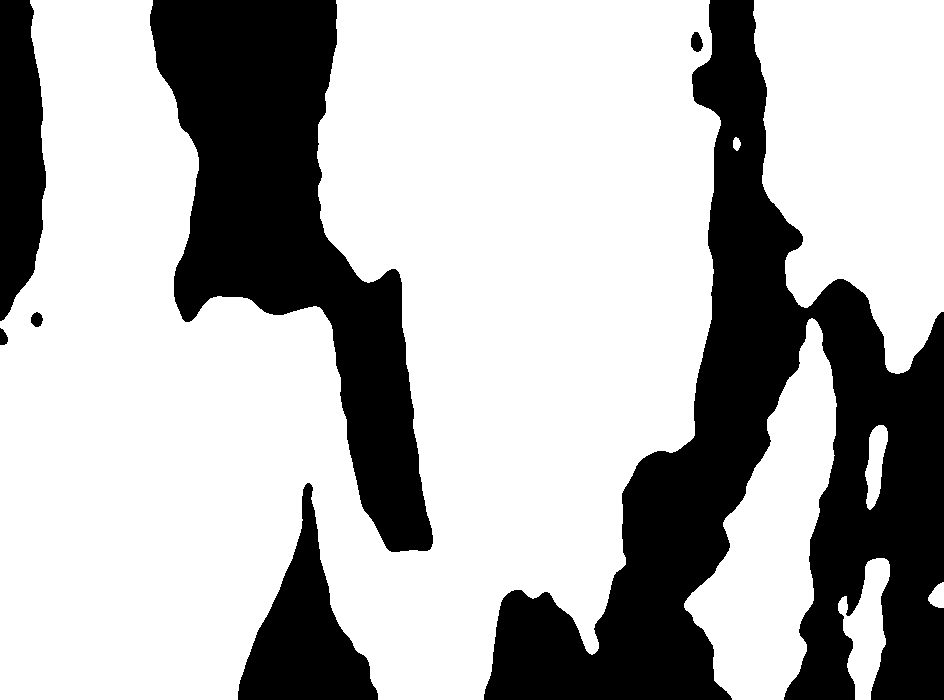}
		\\
	    {\sf \footnotesize Iteration no: 3} & {\sf \footnotesize Iteration no: 100} & {\sf \footnotesize Iteration no: 200}  & {\sf \footnotesize Iteration no: 434}\\

	\end{tabular}
	\caption{Snapshots of the evolving segmentation $\{u(x)>0.5\}$ from intermediate iterations.}
	\label{fig:snapshots}
\end{figure}

\begin{figure}[!htbp]
    \begin{center}
	\begin{tabular}{ccc}
		\includegraphics[width=0.31\linewidth]{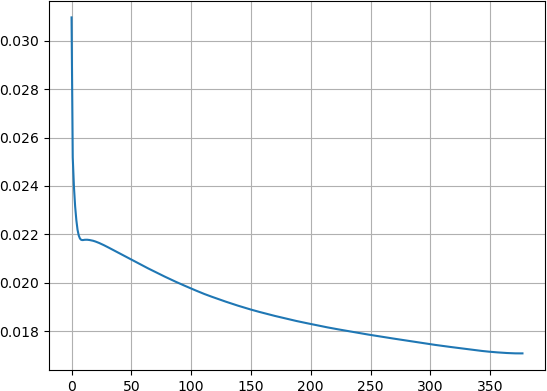} &
		\includegraphics[width=0.31\linewidth]{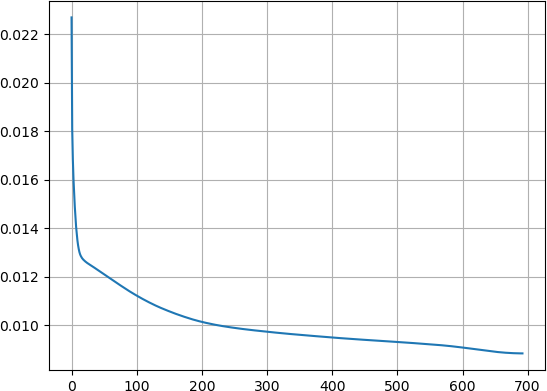} &
		\includegraphics[width=0.31\linewidth]{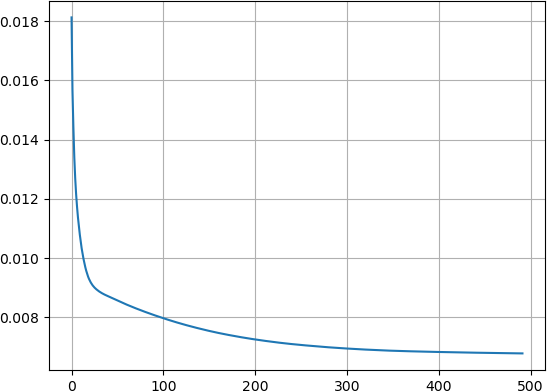} \\

		\includegraphics[width=0.31\linewidth]{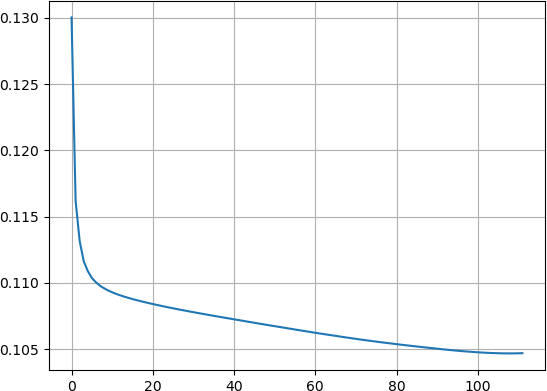} &
		\includegraphics[width=0.31\linewidth]{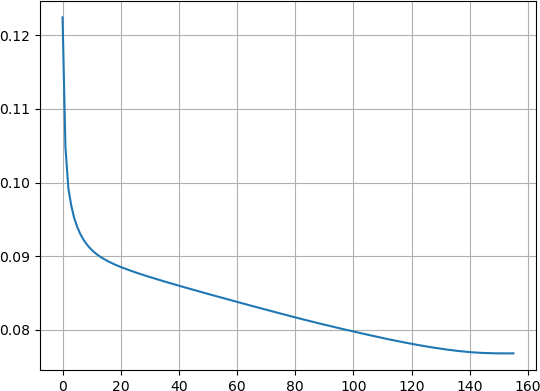} &
		\includegraphics[width=0.31\linewidth]{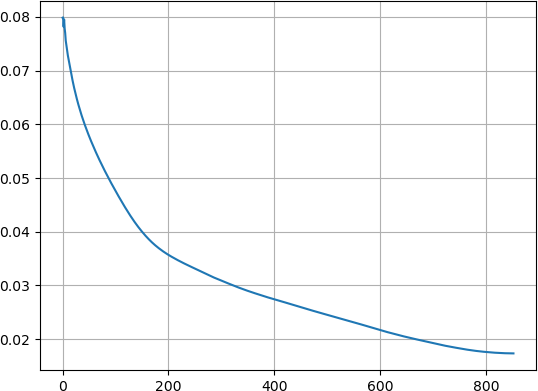} \\

		\includegraphics[width=0.31\linewidth]{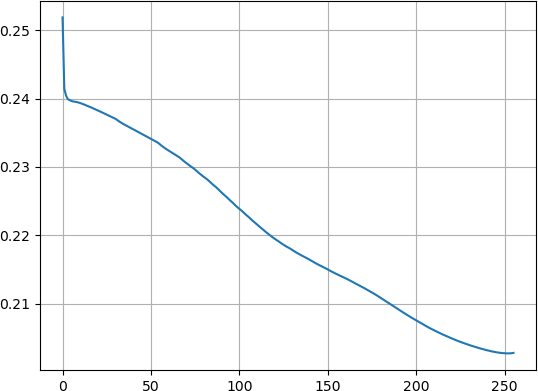} &
		\includegraphics[width=0.31\linewidth]{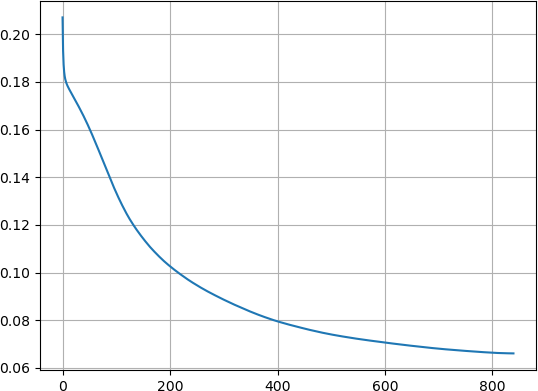} &
		\includegraphics[width=0.31\linewidth]{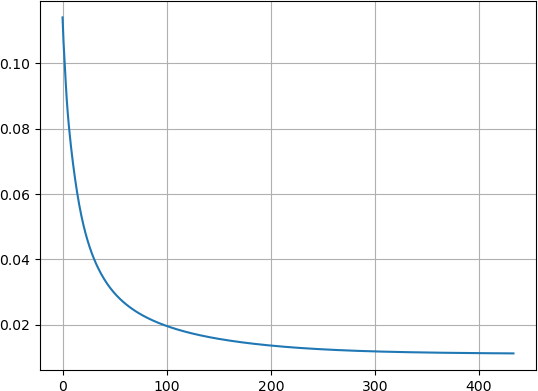} \\

        \end{tabular}
	\caption{Evolution of the energy values
	for galaxy (top row), microstructure (middle row), and pearlite (bottom row) images in the cases of low (left column), medium (middle column) and high (right column) regularization.}
	\label{fig:energy-udiff-tau}
    \end{center}
\end{figure}

\paragraph{Observing the iterations.}
We observed the evolution of the function $u$ and the change in the energy for the three example images.
In Figure~\ref{fig:snapshots}, we illustrate the evolution of the segmentation $\{u(x)>0.5\}$ by taking some snapshots from intermediate iterations of minimization as it converges towards the optimal solution. 
Looking closely at the galaxy example, we observe that the function $u$ is still very close to the initial value $f$ at iteration $k=5$, and its energy is still high as seen in Figure~\ref{fig:energy-udiff-tau}. After 20 iterations, the function $u$ starts to take the shape of our optimal solution. This can be observed  from the energy plot as well. The energy drops sharply after about 30 to 50 more iterations (see first row of Figure~\ref{fig:energy-udiff-tau}). 

After the initial rapid progress, the updates between subsequent iterations are smaller. For example, in the evolution of $u$ for the galaxy image, the changes between the last few hundred iterations are small. This is also observed in some of the other energy plots in Figure~\ref{fig:energy-udiff-tau} where we see in some cases the energy steadily decreasing at a very slow rate. The same behavior can be observed for the other test images as shown in Figure~\ref{fig:snapshots} and Figure~\ref{fig:energy-udiff-tau}. 

\clearpage

\begin{figure}[!htb]
	\begin{tabular}{ccc}
		{\sf \footnotesize low regularization} & {\sf \footnotesize medium regularization} & {\sf \footnotesize high regularization}  \\
		\includegraphics[width=0.30\linewidth]{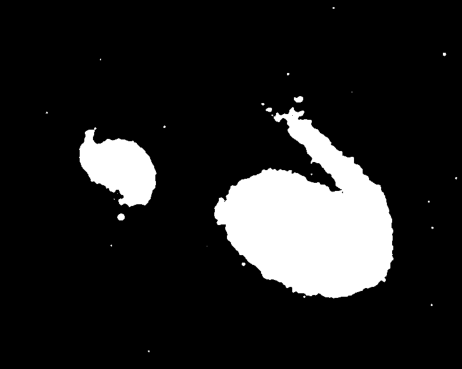} &
		\includegraphics[width=0.30\linewidth]{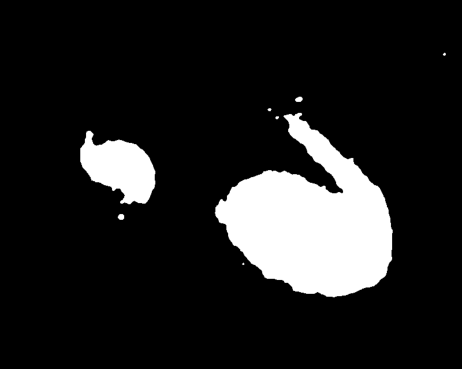}  &
		\includegraphics[width=0.30\linewidth]{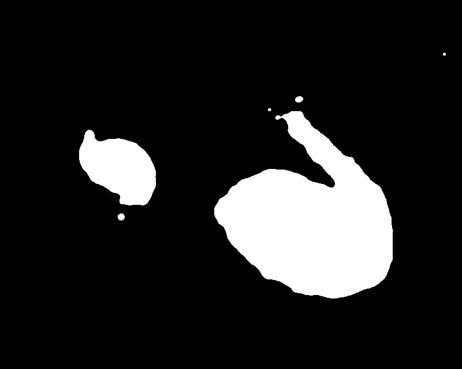} \\

		{\sf \footnotesize $\lambda=5.0$,  iters: 378, time: 21s} &
		{\sf \footnotesize $\lambda=2.0$,  iters: 693, time: 35s} &
		{\sf \footnotesize $\lambda=1.0$,  iters: 492, time: 24s} \\

        \includegraphics[width=0.30\linewidth]{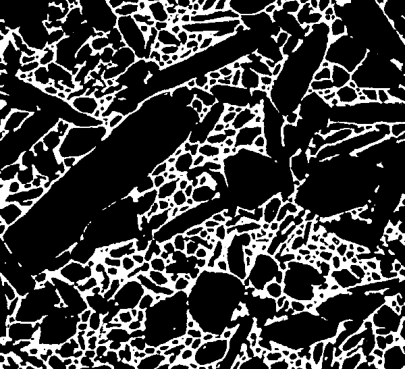} &
	    \includegraphics[width=0.30\linewidth]     {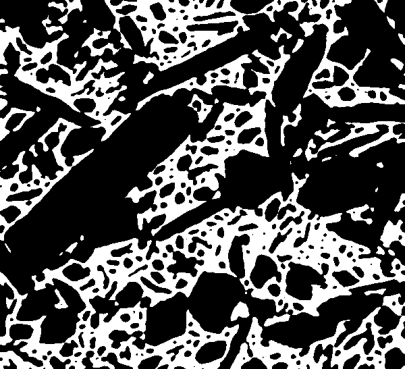}  &
	    \includegraphics[width=0.30\linewidth]{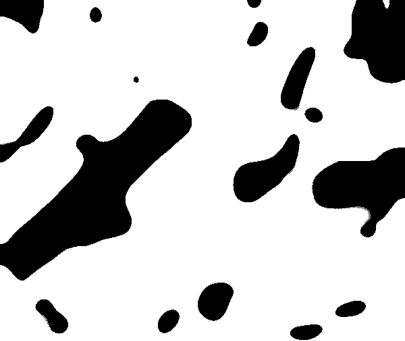} \\

		{\sf \footnotesize $\lambda=2.5$,  iters: 109, time: 3.9s} &
		{\sf \footnotesize $\lambda=0.5$,  iters: 156, time: 5.1s} &
		{\sf \footnotesize $\lambda=0.1$,  iters: 853, time: 31s} \\

		\includegraphics[width=0.30\linewidth]{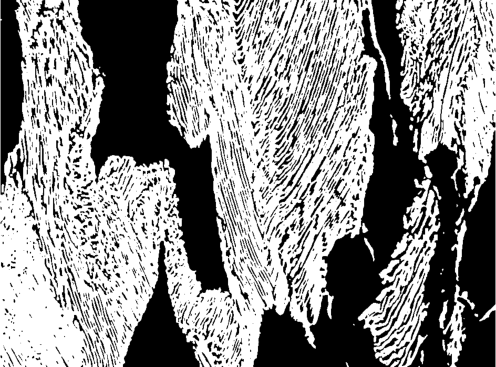} &
		\includegraphics[width=0.30\linewidth]{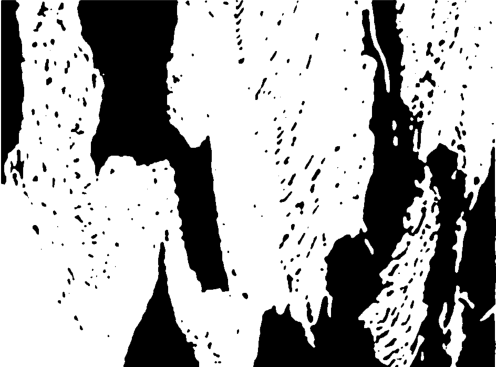}  &
		\includegraphics[width=0.30\linewidth]{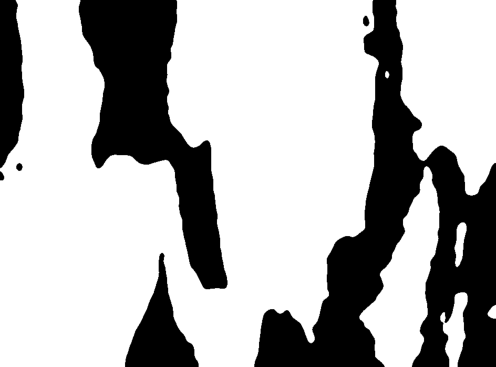} \\

		{\sf \footnotesize $\lambda=5.0$,  iters: 256, time: 18s} &
		{\sf \footnotesize $\lambda=1.5$,  iters: 840, time: 61s} &
		{\sf \footnotesize $\lambda=0.1$,  iters: 434, time: 32s} \\

	\end{tabular}
	\caption{Higher values of $\lambda$ result in reduced regularization of 
	region boundaries (left), whereas lower values lead to increased regularization, 
	therefore, smoother region boundaries (middle, right).}
	\label{fig:lambda-tests}
\end{figure}

\paragraph{The effect of data weight.}
We also considered the effects of the data weights $\lambda$ on the segmentation results (see Figure~\ref{fig:lambda-tests}). For a higher data weight $\lambda$, we observed lower regularization in segmentation. In this situation, the regions had rougher boundaries, and the segmentation seemed less smooth. We then considered a medium value of the data weight, and we observed a segmentation with better regularization than in our previous observation, but still with some rough boundaries. 
We then gradually decreased the data weight $\lambda$, and this resulted in increased regularization and smoother region boundaries. 
If we decreased $\lambda$ very much, then the smoothing was excessive and destroyed many details in the segmentation, as shown in the middle row of Figure~\ref{fig:lambda-tests} for data weight $\lambda=0.1$. However, the same data weight resulted in a desirable segmentation for the pearlite image (bottom row of Figure~\ref{fig:lambda-tests}), as the strong smoothing closed the gaps in the bright regions.

We did not observe a direct correlation between the value $\lambda$ and the number of iterations or computation time. The number of iterations in each experiment rather reflected the amount of work needed to reach the minimum of the energy, depending on how far the initial iterate $u^0$ was from the final segmentation.
The details of the data weight experiments for the three test images are shown in Figure~\ref{fig:lambda-tests}.

\paragraph{Comparisons with other algorithms.}
Finally, we compared the segmentations from our implementation to those obtained using two closely related algorithms: Otsu thresholding~\cite{otsu,balarini:nes} and Chan-Vese segmentation~\cite{chan:vese,getreuer:CV}. Otsu thresholding groups the image pixels into two regions, based on the pixel values and whether the values are higher or lower than a threshold value. The threshold value can be selected automatically by the Otsu algorithm, which is the option we used in our experiments with the implementation of~\cite{balarini:nes}. The Chan-Vese model~\cite{chan:vese} is essentially the same energy minimized in this paper (which follows~\cite{gold:bre}). In~\cite{chan:vese}, the segmentation energy is minimized by the gradient descent evolution of a level set function representing the domains. For this computation, we used the C++ implementation of~\cite{getreuer:CV}. This implementation had multiple parameters, such as the length penalty weight $\mu$, which we set to 0.2 for galaxy and 0.1 for microstructure and pearlite images, other model parameters (for which we used the defaults) of $\nu=0$ (area weight), $\lambda_1=1$ (fit weight of the foreground region), $\lambda_2=1$ (fit weight of the background region), and the iteration parameters were set to $dt=1.0$, $maxiter=1000$, $tol=0.001$ for the galaxy and $tol=0.002$ for the microstructure and pearlite images.

\begin{figure}[!htb]
	\begin{tabular}{ccc}
		  {\sf \footnotesize Bregman iterations} & {\sf \footnotesize Otsu thresholding} & {\sf footnotesize Chan-Vese level sets}  \\

		\includegraphics[width=0.30\linewidth]{galaxy_lambda_200.png} &
		\includegraphics[width=0.30\linewidth]{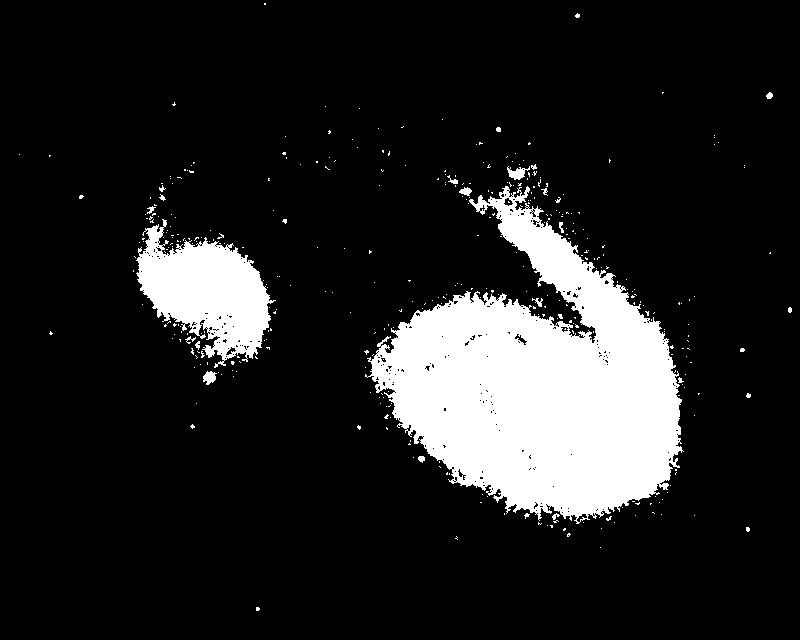}  &
		\includegraphics[width=0.30\linewidth]{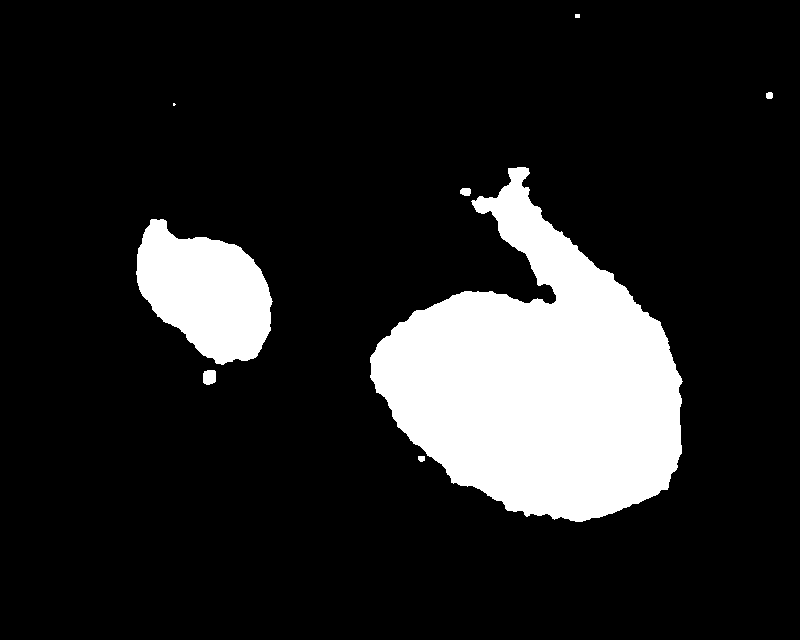} \\

		{\sf \footnotesize data weight $\lambda=2.0$} &
		{\sf \footnotesize auto-threshold} &
		{\sf \footnotesize length penalty $\mu=0.1$} \\

        \includegraphics[width=0.30\linewidth]{microstructure_lambda_050.png} &
	    \includegraphics[width=0.30\linewidth]{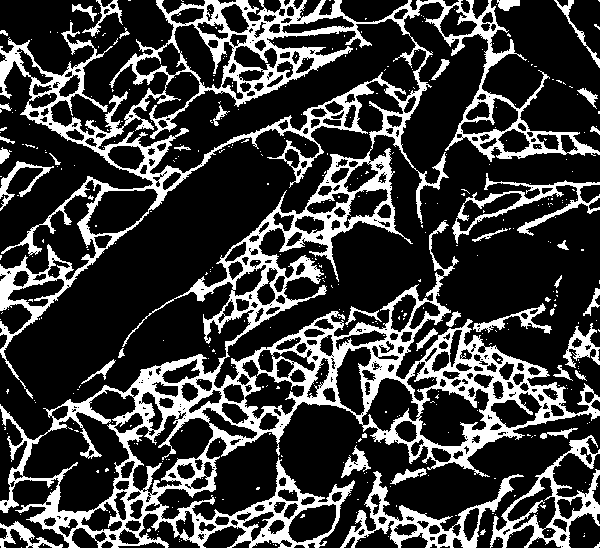}  &
	    \includegraphics[width=0.30\linewidth]{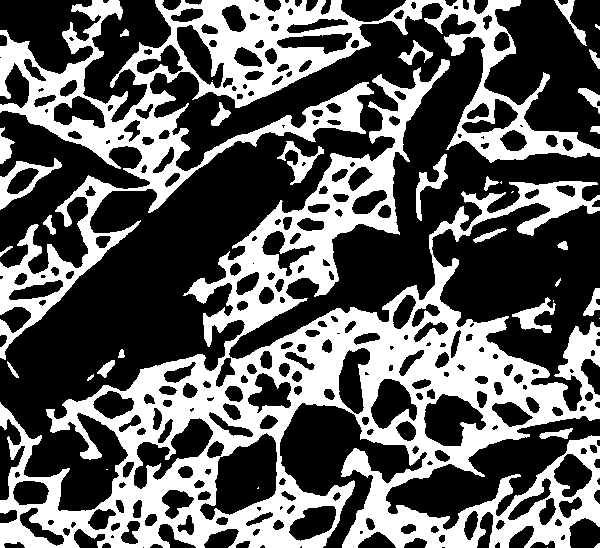} \\

		{\sf \footnotesize data weight $\lambda=0.5$} &
		{\sf \footnotesize auto-threshold} &
		{\sf \footnotesize length penalty $\mu=0.2$} \\

		\includegraphics[width=0.30\linewidth]{pearlite_lambda_150.png} &
		\includegraphics[width=0.30\linewidth]{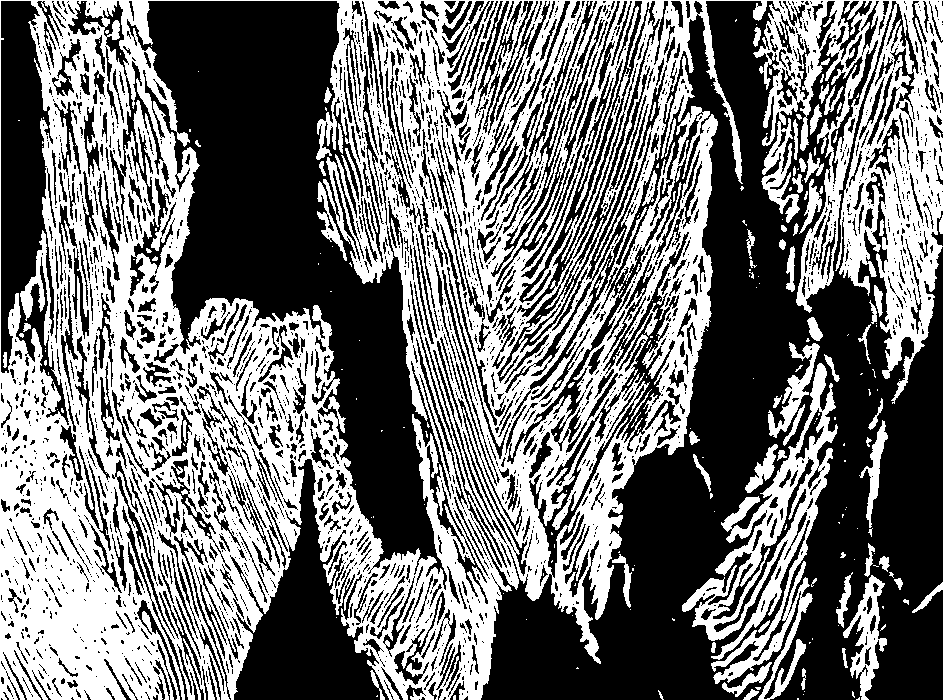}  &
		\includegraphics[width=0.30\linewidth]{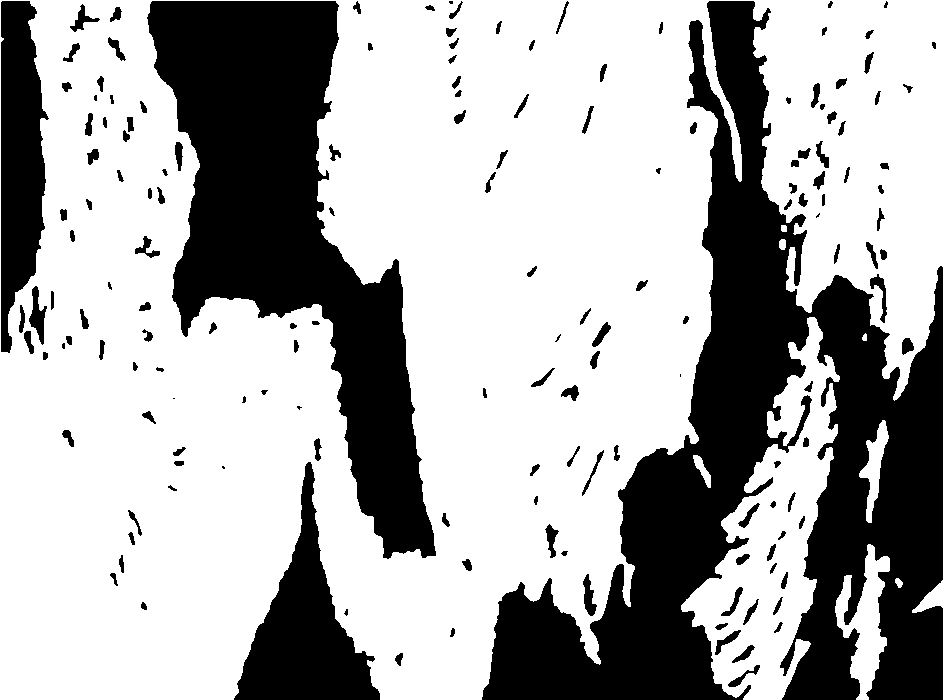} \\

		{\sf \footnotesize data weight $\lambda=1.5$} &
		{\sf \footnotesize auto-threshold} &
		{\sf \footnotesize length penalty $\mu=0.2$} \\

	\end{tabular}
	\caption{Segmentation results obtained using Bregman iterations (left),
        Otsu thresholding with auto-threshold (middle), 
        Chan-Vese level set method (right).
        Bregman iterations and Chan-Vese algorithms produce comparable
        smooth segmentations, whereas Otsu thresholding segmentation
        is noisy.
        }
        \label{fig:algorithm-comparisons}
\end{figure}

\begin{figure}[!htbp]
\sf \footnotesize
\centering
	\begin{tabular}{cccc}

		  Input image & Bregman iterations & Otsu thresholding & Chan-Vese level sets  \\ \\

		\includegraphics[width=0.16\linewidth]{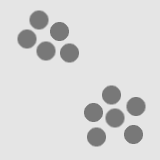} &
		\includegraphics[width=0.16\linewidth]{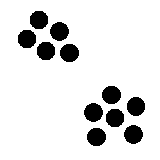} &
		\includegraphics[width=0.16\linewidth]{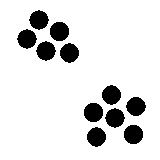} &
		\includegraphics[width=0.16\linewidth]{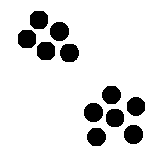}  \\

        Clusters of circles &
		data weight $\lambda=1.0$ &
		auto-threshold &
		length penalty $\mu=0.1$ \\

		\includegraphics[width=0.16\linewidth]{cluster.png} &
		\includegraphics[width=0.16\linewidth]{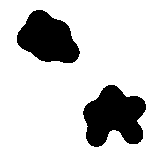} &
	    \includegraphics[width=0.16\linewidth]{cluster_otsu.png} &
		\includegraphics[width=0.16\linewidth]{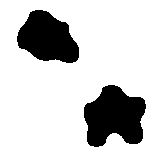}  \\

        Clusters of circles &
		data weight $\lambda=0.085$ &
		auto-threshold &
		length penalty $\mu=0.6$ \\

        \includegraphics[width=0.16\linewidth]{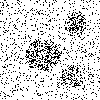} &
        \includegraphics[width=0.16\linewidth]{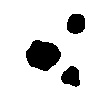} &
	    \includegraphics[width=0.16\linewidth]{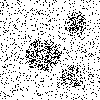} &
	    \includegraphics[width=0.16\linewidth]{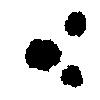}  \\

        Noisy clusters &
		data weight $\lambda=0.25$ &
		auto-threshold &
		length penalty $\mu=0.2$ \\

		\includegraphics[width=0.18\linewidth]{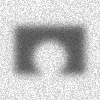} &
		\includegraphics[width=0.18\linewidth]{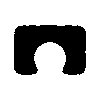} &
		\includegraphics[width=0.18\linewidth]{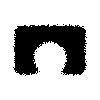}  &
		\includegraphics[width=0.18\linewidth]{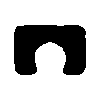} \\

        Blurry nonconvex &
		data weight $\lambda=2.0$ &
		auto-threshold &
		length penalty $\mu=0.2$ \\

		\includegraphics[width=0.18\linewidth]{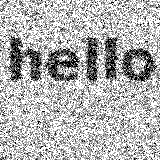} &
		\includegraphics[width=0.18\linewidth]{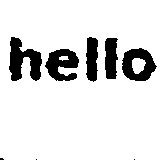} &
		\includegraphics[width=0.18\linewidth]{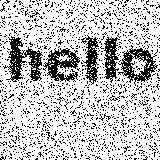}  &
		\includegraphics[width=0.18\linewidth]{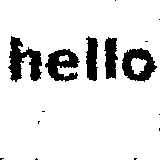} \\

        Noisy hello &
		data weight $\lambda=1.0$ &
		auto-threshold &
		length penalty $\mu=0.2$ \\

		\includegraphics[width=0.18\linewidth]{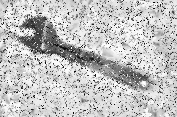} &
		\includegraphics[width=0.18\linewidth]{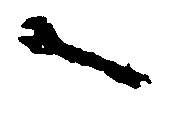} &
		\includegraphics[width=0.18\linewidth]{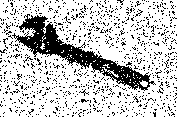}  &
		\includegraphics[width=0.18\linewidth]{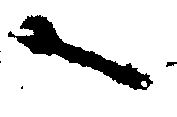} \\

        Noisy wrench &
		data weight $\lambda=2.0$ &
		auto-threshold &
		length penalty $\mu=0.2$ \\

	\end{tabular}
	\caption{Segmentation results on synthetic images from~\cite{getreuer:CV,balarini:nes} (1st column) 
        obtained using Bregman iterations (2nd column),
        Otsu thresholding with auto-threshold (3rd column), 
        Chan-Vese level set method (4th column).
        Bregman iterations and Chan-Vese algorithms produce comparable
        smooth segmentations, whereas Otsu thresholding segmentation
        is noisy.
        }
        \label{fig:more-comparisons-1}
\end{figure}

\begin{figure}[!htbp]
\sf \footnotesize
\centering
	\begin{tabular}{cccc}

		  Input image & Bregman iterations & Otsu thresholding & Chan-Vese level sets  \\ \\

		\includegraphics[width=0.2\linewidth]{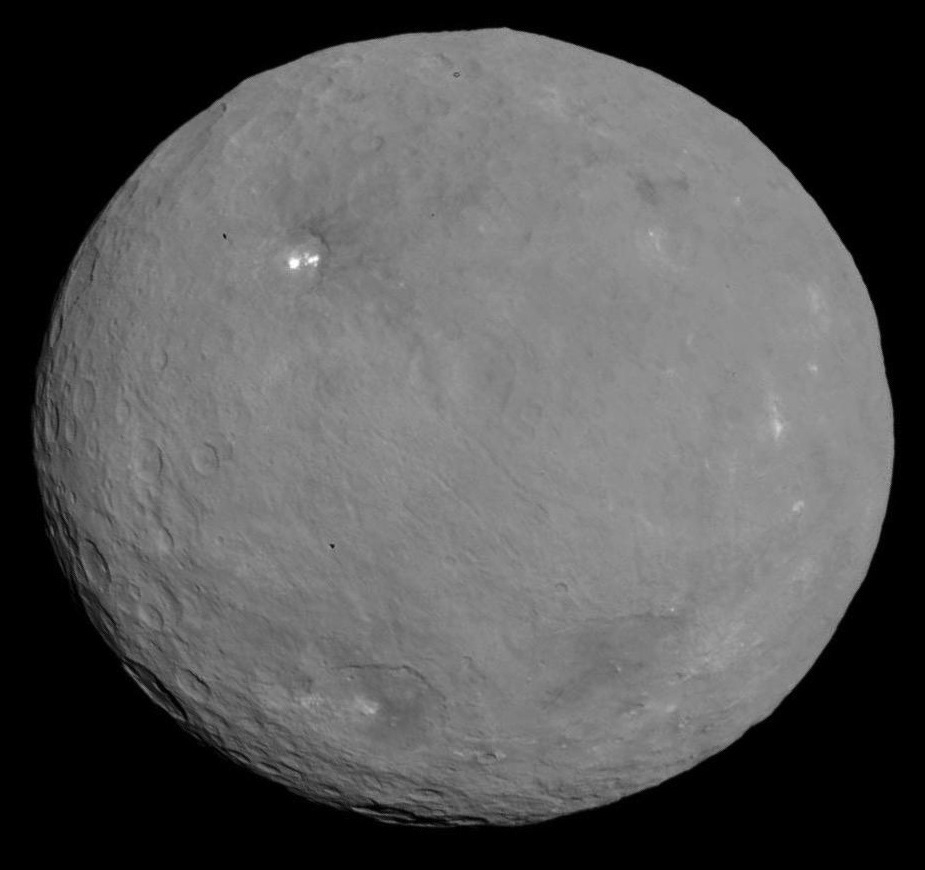} &
		\includegraphics[width=0.2\linewidth]{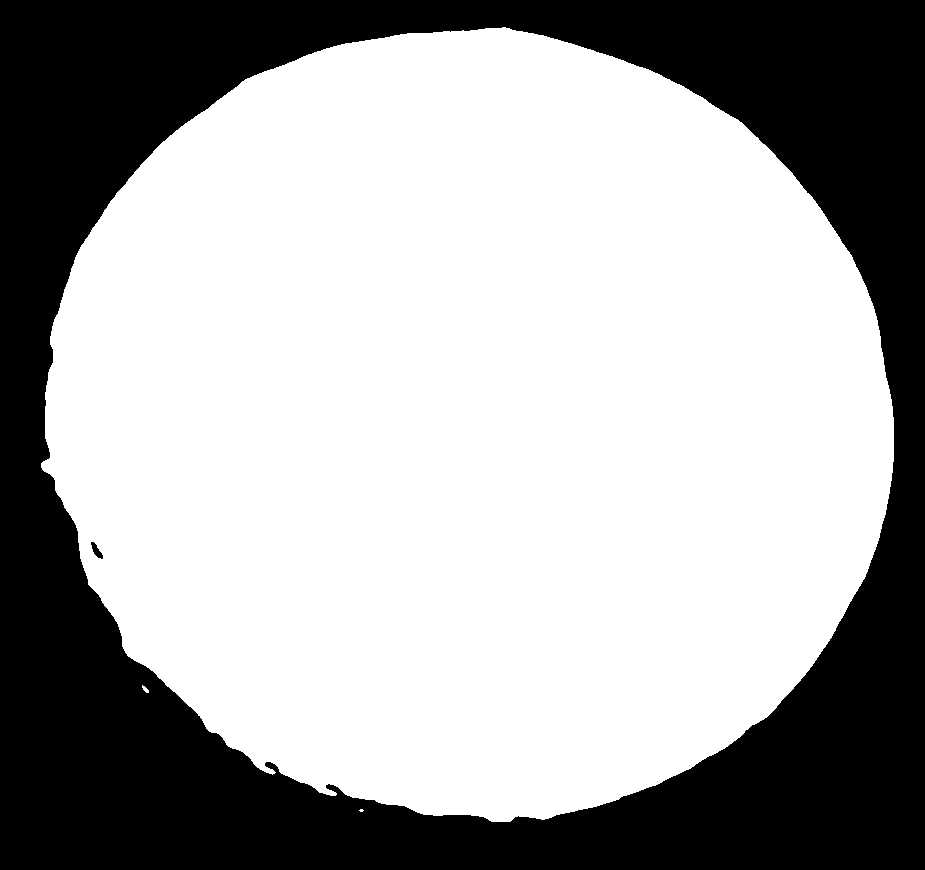} &
		\includegraphics[width=0.2\linewidth]{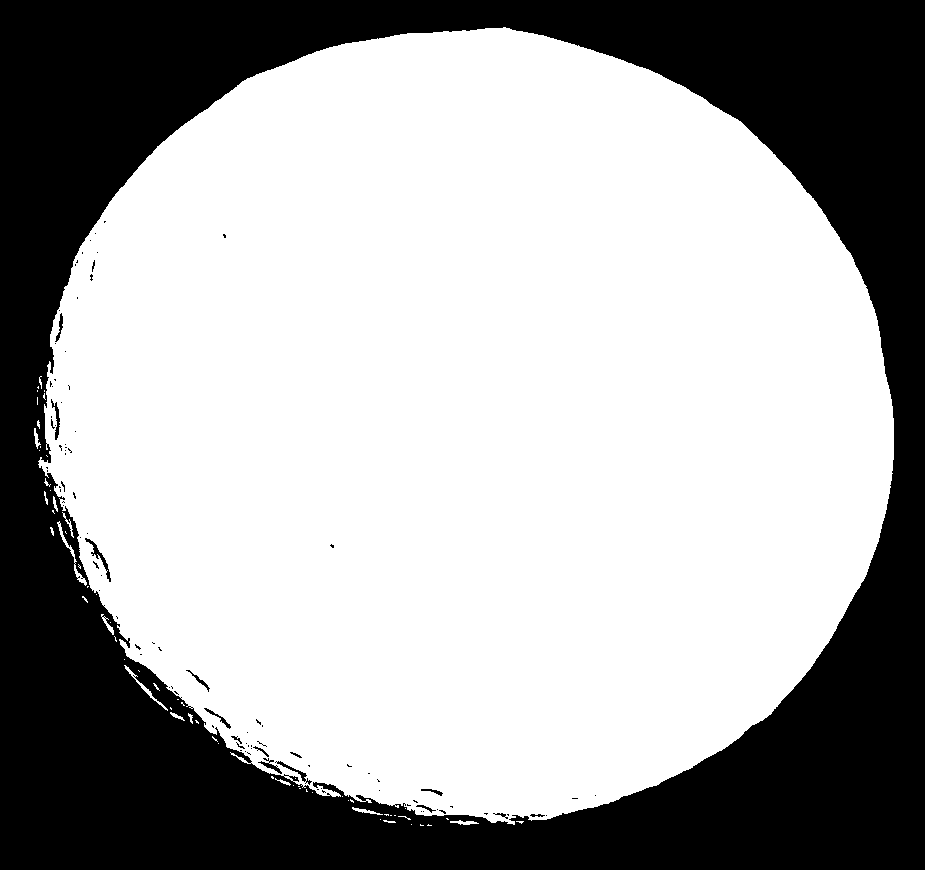} &
		\includegraphics[width=0.2\linewidth]{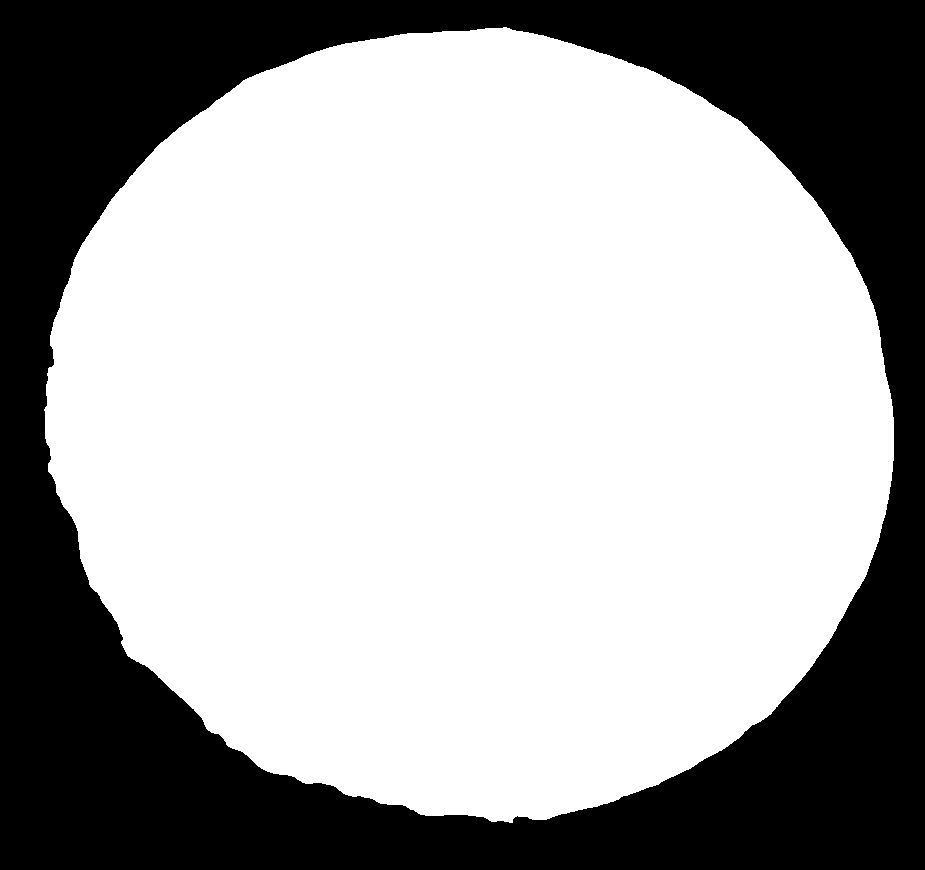}  \\

        Ceres &
		data weight $\lambda=2.0$ &
		auto-threshold &
		length penalty $\mu=0.2$ \\

		\includegraphics[width=0.23\linewidth]{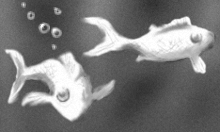} &
		\includegraphics[width=0.23\linewidth]{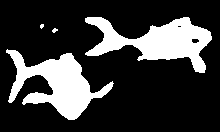} &
	    \includegraphics[width=0.23\linewidth]{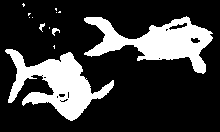} &
		\includegraphics[width=0.23\linewidth]{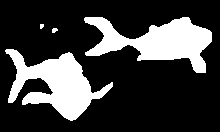}  \\

        Fish &
		data weight $\lambda=2.0$ &
		auto-threshold &
		length penalty $\mu=0.2$ \\

        \includegraphics[width=0.23\linewidth]{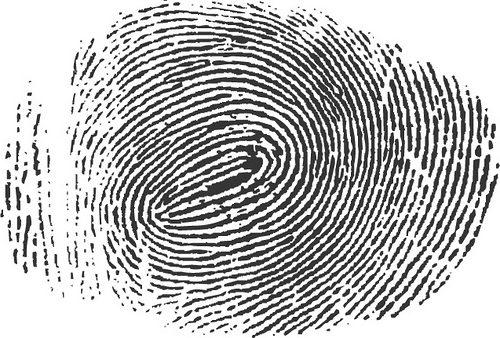} &
        \includegraphics[width=0.23\linewidth]{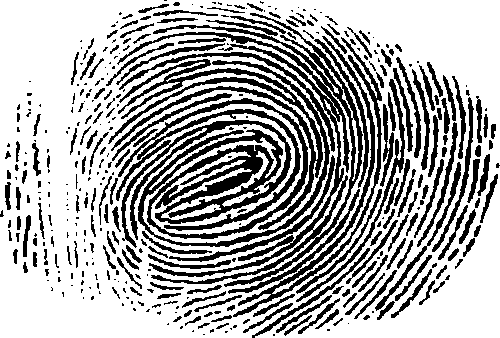} &
	    \includegraphics[width=0.23\linewidth]{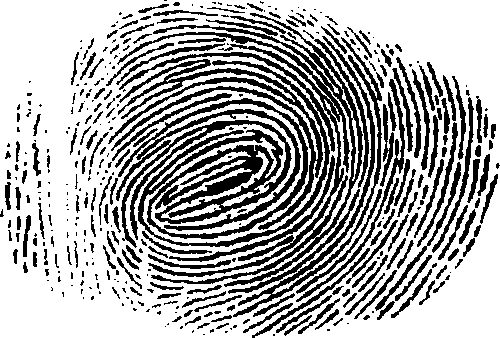} &
	    \includegraphics[width=0.23\linewidth]{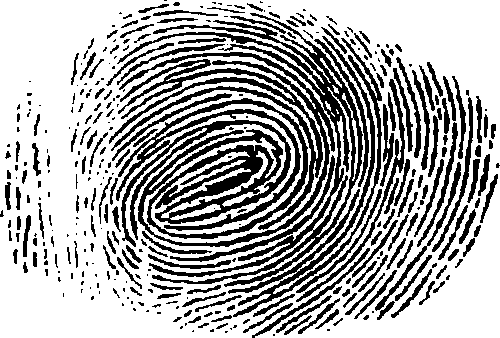}  \\

        Fingerprint &
		data weight $\lambda=2.0$ &
		auto-threshold &
		length penalty $\mu=0.2$ \\

		\includegraphics[width=0.23\linewidth]{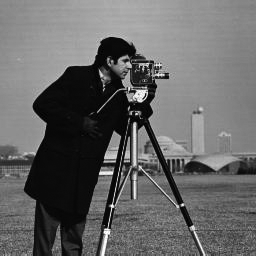} &
		\includegraphics[width=0.23\linewidth]{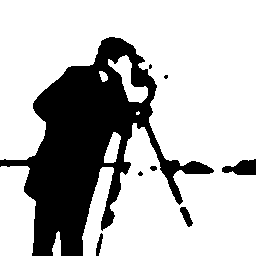} &
		\includegraphics[width=0.23\linewidth]{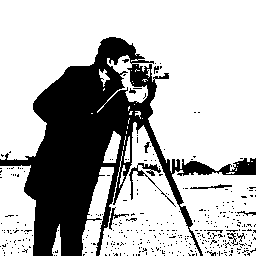}  &
		\includegraphics[width=0.23\linewidth]{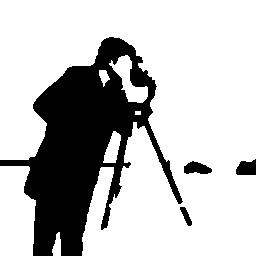} \\

        Cameraman &
		data weight $\lambda=2.0$ &
		auto-threshold &
		length penalty $\mu=0.2$ \\

		\includegraphics[width=0.23\linewidth]{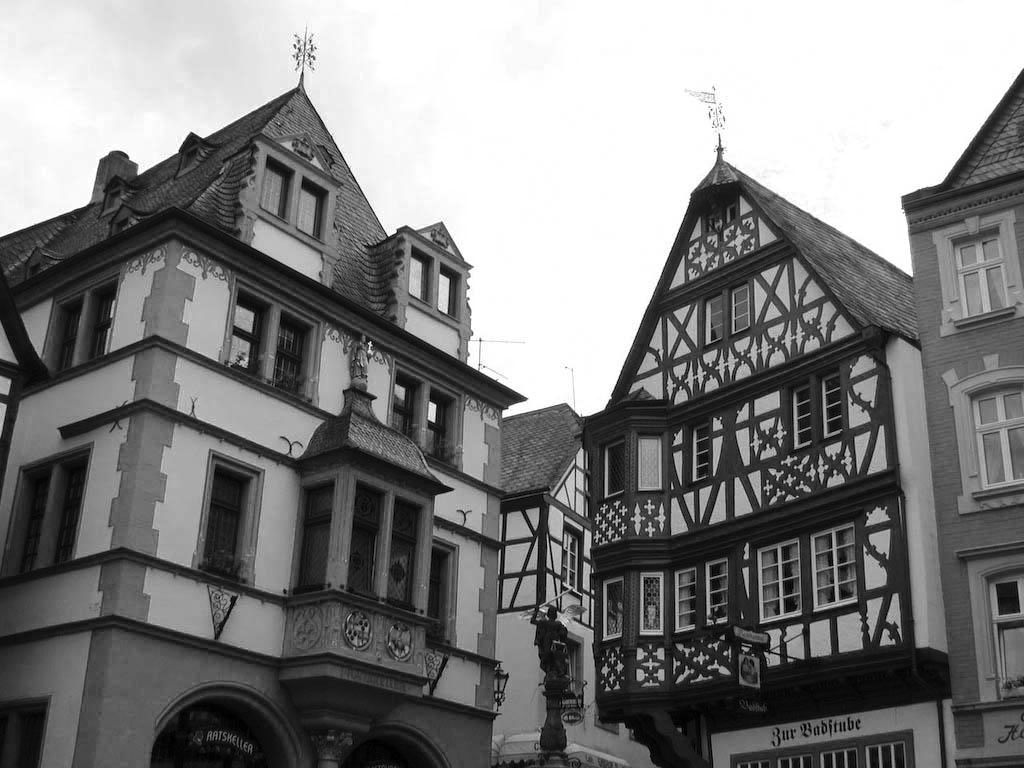} &
		\includegraphics[width=0.23\linewidth]{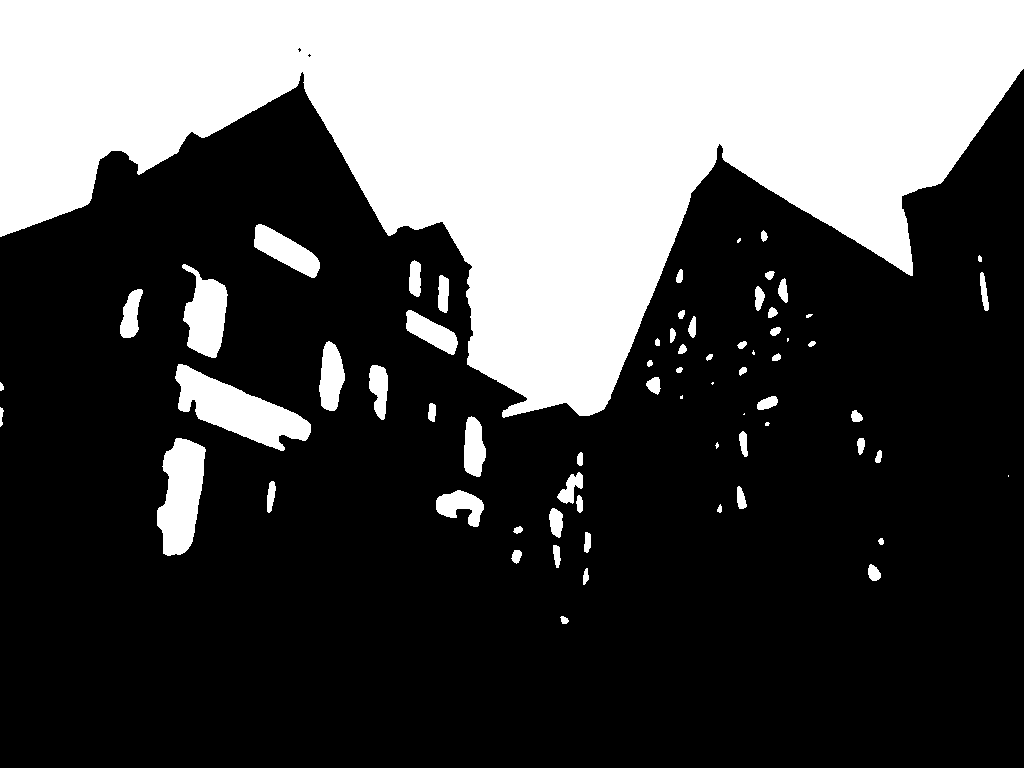} &
		\includegraphics[width=0.23\linewidth]{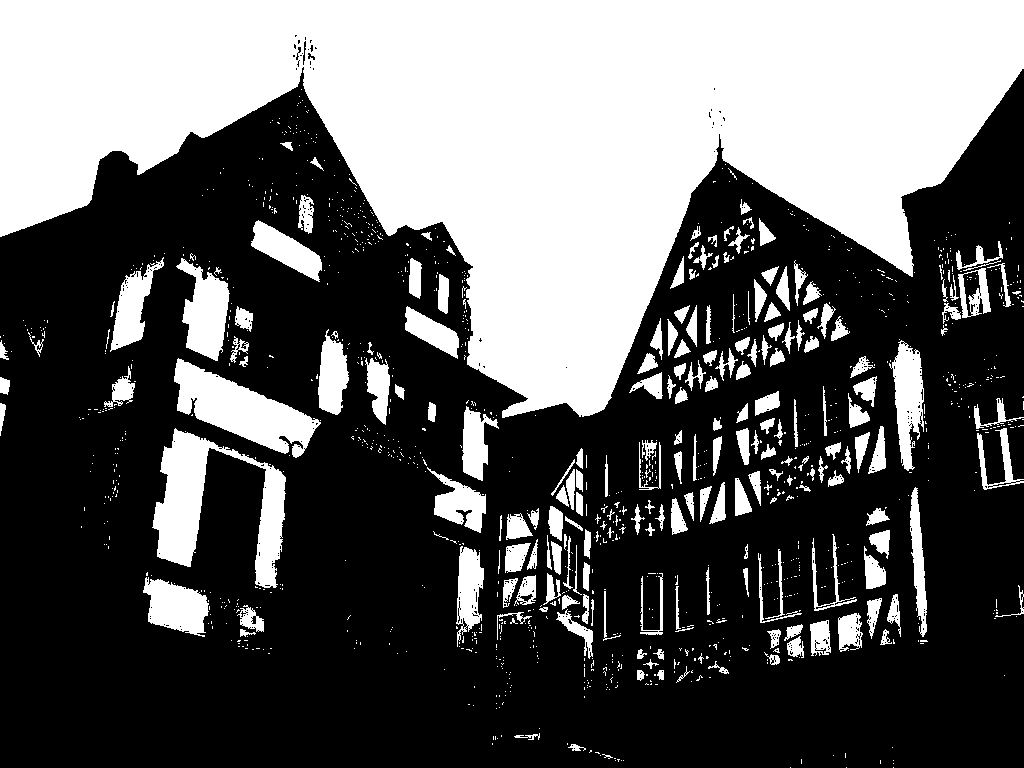}  &
		\includegraphics[width=0.23\linewidth]{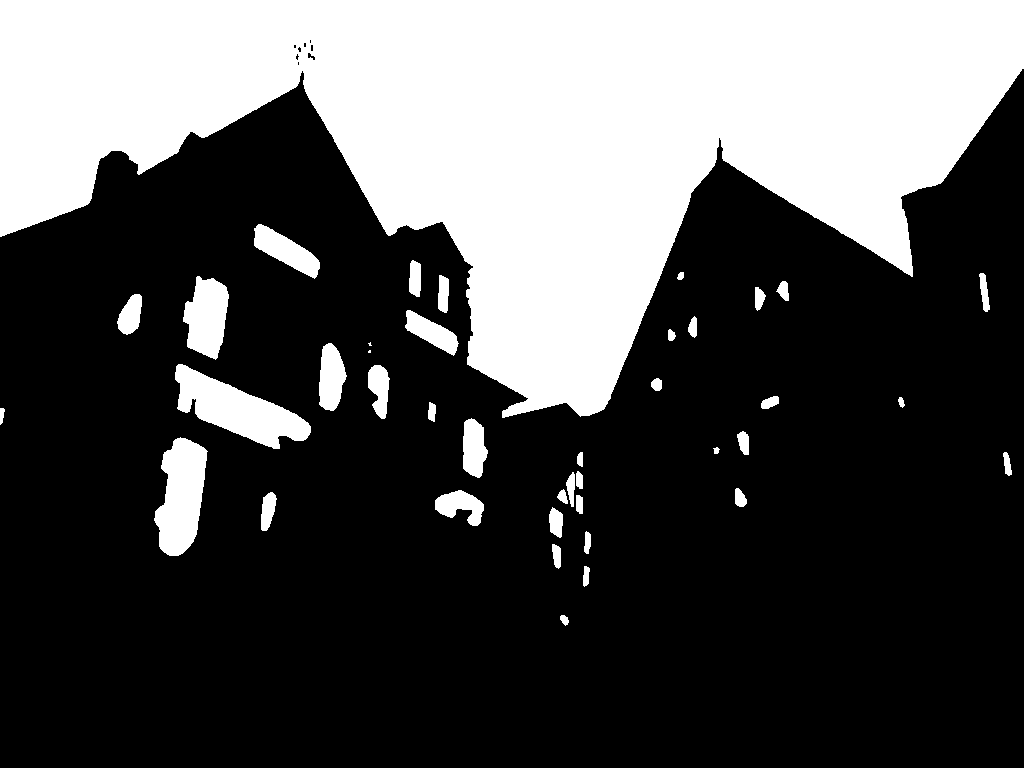} \\

        Houses &
		data weight $\lambda=2.0$ &
		auto-threshold &
		length penalty $\mu=0.2$ \\

	\end{tabular}
	\caption{Segmentation results on real images from~\cite{getreuer:CV,balarini:nes} (1st column) 
        obtained using Bregman iterations (2nd column),
        Otsu thresholding with auto-threshold (3rd column), 
        Chan-Vese level set method (4th column).
        }
        \label{fig:more-comparisons-2}
\end{figure}

The segmentations obtained using these three methods are shown in Figure~\ref{fig:algorithm-comparisons}. As expected, the segmentation results from Otsu thresholding are noisy, because it does not have boundary or region regularization. The final results of the Chan-Vese segmentation are comparable to the results from our implementation of Bregman iterations. We include additional comparisons using images from~\cite{getreuer:CV,balarini:nes} in Figures~\ref{fig:more-comparisons-1} and~\ref{fig:more-comparisons-2}. Note that the $\mu$ weight of the boundary length penalty in the Chan-Vese model, as well as our $\lambda$ data weight, can be adjusted to obtain smoother (or less smooth) segmentations. The level set minimization of the Chan-Vese segmentation using~\cite{getreuer:CV} took many more iterations than our Bregman algorithm solutions (which had motivated this later approach). We should still note that the computational cost and the number of iterations of the Chan-Vese algorithm depend very much on the values of $tol$, $maxiter$, and $dt$, so it can be considerably higher or lower. In any case, the C++ implementation was very efficient and fast.

\clearpage
\section*{Disclaimer} 
Certain equipment, instruments, software, or materials, commercial or non-commercial, are identified in this paper in order to specify the experimental procedure adequately. Such identification is not intended to imply recommendation or endorsement of any product or service by NIST, nor is it intended to imply that the materials or equipment identified are necessarily the best available for the purpose.

\section*{Acknowledgments}
The first author was partially supported by the Mathematical Sciences Graduate Internship (MSGI) Program of U.S. National Science Foundation (NSF) during his internship at the U.S. National Institute of Standards and Technology (NIST), also by the Alexander von Humboldt Foundation German Research Chair funding and the associated Der Deutsche Akademische Austauschdienst (DAAD) Projects No. 57610033 and 57761435.

\section*{Image Credits}
\small
\includegraphics[height=2em]{galaxy.png} Wikipedia\footnote{\url{https://de.m.wikipedia.org/wiki/NGC_5278/79}},\\
\includegraphics[height=2em]{microstructure.png} Figure 6 in~\cite{swab2005mechanical},\\
\includegraphics[height=2em]{pearlite.png} Wikimedia Commons\footnote{\url{https://upload.wikimedia.org/wikipedia/commons/7/77/Pearlite.jpg}},\\
\includegraphics[height=2em]{cluster.png}
\includegraphics[height=2em]{noisy_cluster.png}
\includegraphics[height=2em]{nonconvex_blurry.png}
\includegraphics[height=2em]{hello.png}
\includegraphics[height=2em]{fish.png}
\includegraphics[height=2em]{wrench.png} Pascal Getreuer~\cite{getreuer:CV}, CC-BY,\\
\includegraphics[height=2em]{ceres.png}
  Ceres Dwarf Planet (NASA/JPL-Caltech/UCLA/MPS/DLR/IDA), Public Domain\footnote{\url{http://photojournal.jpl.nasa.gov/jpeg/PIA19562.jpg}},\\
\includegraphics[height=2em]{houses.png}
  Houses (Wikipedia),
  CC-BY-SA\footnote{\url{https://commons.wikimedia.org/wiki/File:Image_processing_pre_otsus_algorithm.jpg}},\\
\includegraphics[height=2em]{fingerprint.png}
Fingerprint (Wikipedia), CC-BY-SA 3.0\footnote{\url{https://commons.wikimedia.org/wiki/File:Fingerprintforcriminologystubs.jpg}},\\ 
\includegraphics[height=2em]{cameraman.png}
  Cameraman, standard test image.

\bibliographystyle{siam}
\bibliography{references}

@article{gold:bre,
  author  = {Goldstein, T. and Bresson, X. and Osher, S.},
  title   = {Geometric Applications of the Split Bregman Method: Segmentation and Surface Reconstruction},
  journal = {J. Sci. Comput.},
  volume  = {45},
  pages   = {272--293},
  year    = {2010},
  note    = {doi:10.1007/s10915-009-9331-z}
}

@article{chan:vese,
  author  = {Chan, T. F. and Vese, L. A.},
  title   = {Active Contours without Edges},
  journal = {IEEE Trans. Image Process.},
  volume  = {10},
  number  = {2},
  pages   = {266--277},
  year    = {2001},
  note    = {doi:10.1109/83.902291}
}

@article{mumford:shah,
  author  = {Mumford, D. and Shah, J.},
  title   = {Optimal Approximations by Piecewise Smooth Functions and Associated Variational Problems},
  journal = {Comm. Pure Appl. Math.},
  volume  = {42},
  number  = {5},
  pages   = {577--685},
  year    = {1989},
  note    = {doi:10.1002/cpa.3160420503}
}

@article{kiefer:sto,
  author  = {Kiefer, L. and Storath, M. and Weinmann, A.},
  title   = {PALMS Image Partitioning --- A New Parallel Algorithm for the Piecewise Affine-Linear Mumford--Shah Model},
  journal = {Image Process. On Line},
  volume  = {10},
  pages   = {124--149},
  year    = {2020},
  note    = {doi:10.5201/ipol.2020.295}
}

@article{getreuer:CV,
  author  = {Getreuer, P.},
  title   = {Chan--Vese Segmentation},
  journal = {Image Process. On Line},
  volume  = {2},
  pages   = {214--224},
  year    = {2012},
  note    = {doi:10.5201/ipol.2012.g-cv}
}

@article{otsu,
  author  = {Otsu, N.},
  title   = {A Threshold Selection Method from Gray-Level Histograms},
  journal = {IEEE Trans. Syst., Man, Cybern.},
  volume  = {9},
  number  = {1},
  pages   = {62--66},
  year    = {1979},
  note    = {doi:10.1109/TSMC.1979.4310076}
}

@article{balarini:nes,
  author  = {Balarini, J. P. and Nesmachnow, S.},
  title   = {A C++ Implementation of Otsu's Image Segmentation Method},
  journal = {Image Process. On Line},
  volume  = {6},
  pages   = {155--164},
  year    = {2016},
  note    = {doi:10.5201/ipol.2016.158}
}

@article{chan:ese,
  author  = {Chan, T. F. and Esedoglu, S. and Nikolova, M.},
  title   = {Algorithms for Finding Global Minimizers of Image Segmentation and Denoising Models},
  journal = {SIAM J. Appl. Math.},
  volume  = {66},
  number  = {5},
  pages   = {1632--1648},
  year    = {2006},
  note    = {doi:10.1137/040615286}
}

@article{rudin:osher,
  author  = {Rudin, L. I. and Osher, S. and Fatemi, E.},
  title   = {Nonlinear Total Variation Based Noise Removal Algorithms},
  journal = {Physica D},
  volume  = {60},
  number  = {1--4},
  pages   = {259--268},
  year    = {1992},
  note    = {doi:10.1016/0167-2789(92)90242-F}
}

@article{cham:a,
  author  = {Chambolle, A.},
  title   = {An Algorithm for Total Variation Minimization and Applications},
  journal = {J. Math. Imaging Vis.},
  volume  = {20},
  pages   = {1--8},
  year    = {2004},
  note    = {doi:10.1023/B:JMIV.0000011325.36760.1e}
}

@article{gstein:sh,
  author  = {Goldstein, T. and Osher, S.},
  title   = {The Split Bregman Method for L1-Regularized Problems},
  journal = {SIAM J. Imaging Sci.},
  volume  = {2},
  pages   = {323--343},
  year    = {2009},
  note    = {doi:10.1137/080725891}
}

@article{getr:p,
  author  = {Getreuer, P.},
  title   = {Rudin--Osher--Fatemi Total Variation Denoising Using Split Bregman},
  journal = {Image Process. On Line},
  volume  = {2},
  pages   = {74--95},
  year    = {2012},
  note    = {doi:10.5201/ipol.2012.g-tvd}
}

@article{more:sury,
  author  = {Moreno, J. C. and Surya Prasath, V. B. and Proenca, H. and Palaniappan, K.},
  title   = {Fast and Globally Convex Multiphase Active Contours for Brain MRI Segmentation},
  journal = {Multiscale Model. Simul.},
  volume  = {4},
  pages   = {2281--2289},
  year    = {2006},
  note    = {doi:10.1016/j.cviu.2014.04.010}
}

@article{osher:burg,
  author  = {Osher, S. and Burger, M. and Goldfarb, D. and Xu, J. and Yin, W.},
  title   = {An Iterative Regularization Method for Total Variation‑Based Image Restoration},
  journal = {Multiscale Model. Simul.},
  volume  = {4},
  pages   = {2281--2289},
  year    = {2006},
  note    = {doi:10.1137/040605412}
}

@article{goldste:li,
  author  = {Goldstein, T. and Li, M. and Yuan, X.},
  title   = {Adaptive Primal‑Dual Splitting Methods for Statistical Learning and Image Processing},
  journal = {Adv. Neural Inf. Process. Syst.},
  volume  = {28},
  pages   = {2089--2097},
  year    = {2015}
}

@article{goldstein:set,
  author  = {Goldstein, T. and O'Donoghue, B. and Setzer, S. and Baraniuk, R.},
  title   = {Fast Alternating Direction Optimization Methods},
  journal = {SIAM J. Imaging Sci.},
  volume  = {7},
  pages   = {1588--1623},
  year    = {2014},
  note    = {doi:10.1137/120896219}
}

@misc{swab2005mechanical,
  author       = {Swab, J. J. and Wereszczak, A. A. and Tice, J. and Caspe, R. and Kraft, R. H. and Adams, J. W.},
  title        = {Mechanical and Thermal Properties of Advanced Ceramics for Gun Barrel Applications},
  howpublished = {Army Research Laboratory Technical Report ARL‑TR‑3417},
  year         = {2005}
}

\end{document}